\documentclass{article}
\usepackage{RR}

\usepackage{amsmath}
\usepackage{amssymb}
\usepackage{amsfonts}
\usepackage{times}
\usepackage[numbers,square,sort]{natbib}
\usepackage{hyperref}
\hypersetup{backref,bookmarks=true,pdfpagemode=Fullscreen,linkcolor=red,colorlinks=true,urlcolor=blue}
\DeclareGraphicsExtensions{.eps,.ps}

\numberwithin{subsection}{section}

\numberwithin{subsubsection}{subsection}

\numberwithin{equation}{section}

\newcommand{\noi}{\noindent}
\newcommand{\non}{\nonumber}
\newcommand{\st}{:\,}
\newcommand{\abs}[1]{\lvert#1\rvert}
\newcommand{\bigabs}[1]{\left\lvert#1\right\lvert}
\newcommand{\pr}[2][]{\text{P}_{#1}(#2)}
\newcommand{\del}{\partial}
\newcommand{\integers}{{\mathbb Z}}

\newcommand{\sircle}{S^{1}}
\newcommand{\reg}{R}
\newcommand{\regcmp}{\bar{\reg}}
\newcommand{\boundop}{\partial}
\newcommand{\bound}{\boundop\reg}
\newcommand{\map}{\rightarrow}
\newcommand{\dom}[1]{\Box #1}
\newcommand{\bra}[1]{\langle #1 |}
\newcommand{\ket}[1]{| #1 \rangle}
\newcommand{\braket}[2]{\langle #1 | #2 \rangle}
\newcommand{\braoket}[3]{\bra{#1} #2 \ket{#3}}
\newcommand{\imdom}{\Omega}
\newcommand{\imag}{I}
\newcommand{\knowledge}{K}
\newcommand{\laspace}{\;}
\newcommand{\mespace}{\:}
\newcommand{\intspace}{\mespace}
\newcommand{\isp}{\intspace}
\newcommand{\eg}{{\em e.g.\ }}
\newcommand{\ie}{{\em i.e.\ }}
\newcommand{\etal}{{\em et al.\ }}
\newcommand{\eqcomma}{\laspace,}
\newcommand{\eqstop}{\laspace.}
\newcommand{\eqsemi}{\laspace;}
\newcommand{\bfvec}[1]{#1}
\newcommand{\bfhat}[1]{\hat{\bfvec{#1}}}
\newcommand{\contour}{\gamma}
\newcommand{\tanvec}{\bfvec{\tau}}
\newcommand{\normvec}{\bfvec{n}}
\newcommand{\normanormvec}{\bfhat{n}}
\newcommand{\normaRvec}{\bfhat{R}}
\newcommand{\curv}{\kappa}
\newcommand{\interactionfunction}{\Phi}
\newcommand{\length}{L}
\newcommand{\area}{A}
\newcommand{\Ei}{E_{\text{i}}}
\newcommand{\Eg}{E_{\text{g}}}
\newcommand{\lambdai}{\lambda_{i}}
\newcommand{\lambdac}{\lambda_{C}}
\newcommand{\alphac}{\alpha_{C}}
\newcommand{\betac}{\beta_{C}}
\newcommand{\talphac}{\tilde{\alpha}_{C}}
\newcommand{\dmin}{d}
\newcommand{\contouro}{\contour_{0}}
\newcommand{\ro}{r_{0}}
\newcommand{\hatro}{\hat{r}_{0}}
\newcommand{\tro}{\tilde{r}_{0}}
\newcommand{\thetao}{\theta_{0}}
\newcommand{\dcontour}{\delta\contour}
\newcommand{\dr}{\delta r}
\newcommand{\Dtheta}{\Delta t}
\newcommand{\imagin}{\imag_{\reg}}
\newcommand{\imagout}{\imag_{\regcmp}}

\newcommand{\muin}{\mu}
\newcommand{\muout}{\bar{\mu}}
\newcommand{\sigmain}{\sigma}
\newcommand{\sigmaout}{\bar{\sigma}}
\newcommand{\timeparam}{s}

\RRdate{November 2006} 
\RRauthor{P{\'e}ter Horv{\'a}th
  \and
  Ian H. Jermyn
  \and
  Zoltan Kato
  \and
  Josiane Zerubia
} 

\authorhead{Horv{\'a}th \& Jermyn \& Kato \& Zerubia} 
\RRtitle{Un mod{\`e}le de contours actifs d'ordre sup{\'e}rieur d'un `gaz de
cercles' et son application {\`a} l'extraction des houppiers} 

\RRetitle{A higher-order active contour model of a `gas of circles'
and its application to tree crown extraction} 
\titlehead{A {HOAC} model of a `gas of circles' and its application to
tree crown extraction}

\RRnote{This work was partially supported by EU project IMAVIS (FP5
IHP-MCHT99), EU project MUSCLE (FP6-507752), Egide PAI Balaton (09020XJ),
OTKA T-046805, and a Janos Bolyai Research Fellowship of HAS. The authors
would like to thank the French National Forest Inventory (IFN) for the
aerial images.} 

\RRresume{Plusieurs probl{\`e}mes en traitement d'image demandent
l'identification de la r{\'e}gion dans le domaine de l'image qui correspond {\`a}
une entit{\'e} donn{\'e}e dans la sc{\`e}ne. La solution automatique de ces probl{\`e}mes
demande des mod{\`e}les qui incorporent une connaissance {\em a priori}
importante de la forme de la r{\'e}gion. Beaucoup de m{\'e}thodes pour l'inclusion
d'une telle connaissance ont des difficult{\'e}s quand la topologie est a
priori inconnue, par exemple quand l'entit{\'e} est compos{\'e}e d'un nombre
inconnu d'objets similaires. Les contours actifs d'ordre sup{\'e}rieur (CAOS)
sont une m{\'e}thode pour la mod{\'e}lisation d'une connaissance a priori sur la
forme non-triviale sans n{\'e}cessairement contraindre la topologie de la
r{\'e}gion, via l'inclusion d'interactions non-locales entre les points du bord
de la r{\'e}gion dans l'{\'e}nergie qui d{\'e}finie le model{\'e}.

Le cas d'un nombre inconnu d'objets circulaires se soul{\`e}ve dans plusieurs
domaines, \eg l'imagerie m{\'e}dicale, biologique, nanotechnologique, et la
t{\'e}l{\'e}d{\'e}tection. Des r{\'e}gions compos{\'e}es d'un nombre inconnu de cercles peuvent
{\^e}tre appel{\'e}es un ``gaz de cercles''. Dans ce rapport, nous pr{\'e}sentons un
model{\'e} CAOS d'un gaz de cercles. Pour garantir des cercles stables, nous
faisons une analyse de stabilit{\'e} via un d{\'e}veloppement de Taylor fonctionnel
autour d'une forme circulaire. Cette analyse fixe un des param{\`e}tres du
mod{\`e}le en fonction des autres, et contraint le reste. En combinaison avec
une {\'e}nergie d'attache aux donn{\'e}es appropri{\'e}e, nous appliquons le mod{\`e}le {\`a}
l'extraction des houppiers dans des images a{\'e}riennes, et montrons que sa
performance est meilleure que d'autres techniques.}

\RRabstract{Many image processing problems involve identifying the region
in the image domain occupied by a given entity in the scene. Automatic
solution of these problems requires models that incorporate significant
prior knowledge about the shape of the region. Many methods for including
such knowledge run into difficulties when the topology of the region is
unknown a priori, for example when the entity is composed of an unknown
number of similar objects. Higher-order active contours (HOACs) represent
one method for the modelling of non-trivial prior knowledge about shape
without necessarily constraining region topology, via the inclusion of
non-local interactions between region boundary points in the energy
defining the model.

The case of an unknown number of circular objects arises in a number of
domains, \eg medical, biological, nanotechnological, and remote sensing
imagery. Regions composed of an a priori unknown number of circles may be
referred to as a `gas of circles'. In this report, we present a HOAC model
of a `gas of circles'. In order to guarantee stable circles, we conduct a
stability analysis via a functional Taylor expansion of the HOAC energy
around a circular shape. This analysis fixes one of the model parameters in
terms of the others and constrains the rest. In conjunction with a suitable
likelihood energy, we apply the model to the extraction of tree crowns from
aerial imagery, and show that the new model outperforms other techniques.}

\RRmotcle{arbre, houppier, extraction, image a{\'e}rienne, ordre sup{\'e}rieur,
contour actif, gaz de cercles, a priori, forme}

\RRkeyword{tree, crown, extraction, aerial image, higher-order,
active contour, gas of circles, prior, shape}

\RRprojet{Ariana} 

\RRtheme{\THCog} 

\URSophia 
\begin{document}

\makeRR   

\tableofcontents

\newpage

\section{Introduction}
\label{sec.introduction}

\begin{figure}[!h]
    \begin{center}
        \begin{tabular}{c}
            \includegraphics[width=0.5\textwidth]{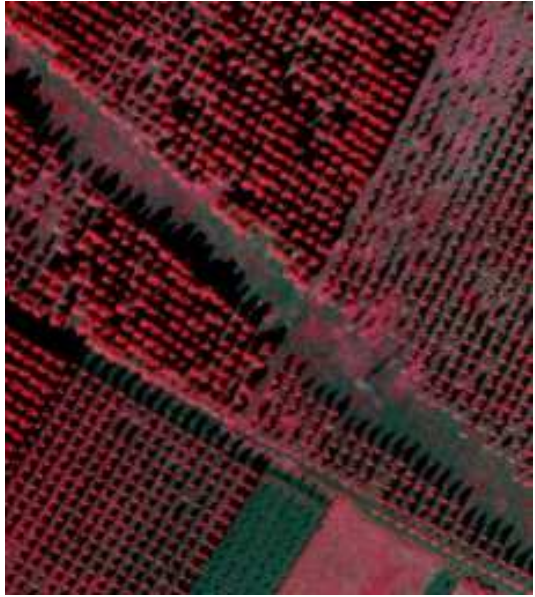}
        \end{tabular}
    \end{center}
    \caption{Real image with planted forest \copyright IFN.}
    \label{fig:real_intro}
\end{figure}

Forestry is a domain in which image processing and computer vision
techniques can have a significant impact. Resource management and
conservation, whether commercial or in the public domain, require
information about the current state of a forest or plantation. Much of this
information can be summarized in statistics related to the size and
placement of individual tree crowns (\eg mean crown area and diameter,
density of the trees). Currently, this information is gathered using
expensive field surveys and time-consuming semi-automatic procedures, with
the result that partial information from a number of chosen sites
frequently has to be extrapolated. An image processing method capable of
automatically extracting tree crowns from high resolution aerial or
satellite images (an example is shown in figure~\ref{fig:real_intro}) and
computing statistics based on the results would greatly aid this domain.

The tree crown extraction problem can be viewed as a special case of a
general image understanding problem: the identification of the region
$\reg$ in the image domain $\imdom$ corresponding to some entity or
entities in the scene. In order to solve this problem in any particular
case, we have to construct, even if only implicitly, a probability
distribution on the space of regions $\pr{\reg | \imag, \knowledge}$. This
distribution depends on the current image data $\imag$ and on any prior
knowledge $\knowledge$ we may have about the region or about its relation
to the image data, as encoded in the likelihood $\pr{\imag | \reg,
\knowledge}$ and the prior $\pr{\reg | \knowledge}$ appearing in the Bayes'
decomposition of $\pr{\reg | \imag, \knowledge}$ (or equivalently in their
energies $-\ln\pr{\imag | \reg, \knowledge}$ and $-\ln\pr{\reg |
\knowledge}$). This probability distribution can then be used to make
estimates of the region we are looking for.

In the automatic solution of realistic problems, the prior knowledge
$\knowledge$, and in particular prior knowledge about the `shape' of the
region, as described by $\pr{\reg | \knowledge}$, is critical. The tree
crown extraction problem provides a good example: particularly in
plantations, $\reg$ takes the form of a collection of approximately
circular connected components of similar size. There is thus a great deal
of prior knowledge about the region sought. The question is then how to
incorporate such prior knowledge into a model for $\reg$? If the model does
not include enough prior knowledge, it will be necessary to include it in
some other form. So, for example, the use of classical active contour
energies to find entities in images usually requires the initialization of
the contour close to the entity to be found, which represents a large
injection of prior knowledge by the user.

The simplest prior information concerns the smoothness properties of the
boundary of the region. For example, the Ising model and many active
contour models~\citep{Kass88,Cohen91,Cohen97,Caselles93,Caselles97a} use
the length of the region boundary and its interior area as their prior
energies. This type of prior information can be augmented using more
demanding measures of smoothness, for example the boundary
curvature~\citep{Kass88,Geman84}. These models are all limited, though, by
the fact that they are integrals over the region boundary of some function
of various derivatives of the boundary. In consequence, they capture local
differential geometric information, corresponding to local interactions
between boundary points, but can say nothing more global about the shape of
the region.

To go further, it is therefore clear that one must introduce longer range
interactions. There are two principal ways to do this: one is to introduce
hidden variables, given which the original variables of interest are (more
or less) independent. Marginalizing over the hidden variables then
introduces interactions between the original variables. Another is to
include explicit long-range interactions between the original variables
(these interactions may also have an interpretation in terms of
marginalization over some hidden variables, which are left implicit).

The first approach has been much investigated, in the form of template
shapes and their deformations
\citep{Chen02,Cremers02,Cremers03a,Cremers03b,Cremers04b,Foulonneau03,Grenander93,Metaxas97,Miller97,Miller02,Paragios02a,Leventon00}.
Here a probability distribution or an energy is defined based on a distance
measure of some kind between regions. One region, the template, is fixed,
while the other is the variable $\reg$. This type of model constrains
$\reg$ to be close to the template region in the space of regions. Template
regions may be learned from examples or fixed by hand; similarly the
distance function maybe based, for example, on the learned covariance of a
Gaussian distribution, or chosen {\em a priori}. The most sophisticated
methods use the kernel trick to define the distance as a pullback from a
high-dimensional space, thereby allowing more complex behaviours. Multiple
templates may also be used, corresponding to a mixture model. It is clear
that these methods implicitly introduce long-range interactions: if you
know that one half of a given region aligns well with the template, this
tells you something about the likely position and shape of the other half.

The above methods assign high probability to regions that lie `close' to
certain points in the space of regions. As such, it is difficult to
construct models of this type that favour regions for which the topology,
and in particular the number of connected components, is unknown {\em a
priori}. There are many problems, however, for which this is the case, for
example, the extraction of networks, or the extraction of an unknown number
of objects of a particular type from astronomical, biological, medical, or
remote sensing images. For this type of prior knowledge, a different type
of model is needed. Higher-order active contours (HOACs) are one such
category of models.

HOACs, first described by~\citet{Rochery03b} (see also
\citep{Rochery05d,Rochery06} for fuller descriptions), take the second
approach mentioned above. They introduce explicit long-range interactions
between region boundary points via energies that contain multiple integrals
over the boundary, thus avoiding the use of template shapes. HOAC energies
can be made intrinsically Euclidean invariant, and, as required by the
above analysis, incorporate sophisticated prior information about region
shape without necessarily constraining the topology of the region. As with
other methods incorporating significant prior knowledge, it is not
necessary to introduce extra knowledge via an initialization close to the
target region: a generic initialization suffices, thus rendering the method
quasi-automatic. \citet{Rochery03b} applied the method to road extraction
from satellite and aerial images using a prior which favours network-like
objects.

In this report, we describe a HOAC model of a `gas of circles': the model
favours regions composed of an {\em a priori} unknown number of circles of
a certain radius. For such a model to work, the circles must be stable to
small perturbations of their boundaries, \ie they must be local minima of
the HOAC energy, for otherwise a circle would tend to `decay' into other
shapes. This is a non-trivial requirement. We impose it by performing a
functional Taylor expansion of the HOAC energy around a circle, and then
demanding that the first order term be zero for all perturbations, and that
the second order term be positive semi-definite. These conditions allow us
to fix one of the model parameters in terms of the others, and constrain
the rest. Experiments using the HOAC energy demonstrate empirically the
coherence between these theoretical considerations and the gradient descent
algorithm used in practice to minimize the energy.

The model has many potential applications, to medical, biological,
physical, and remote sensing imagery in which the entities to be identified
are circular. We choose to apply it to the tree crown extraction problem
from aerial imagery, using the `gas of circles' model as a prior energy,
and an appropriate likelihood. We will see that the extra prior knowledge
included in the `gas of circles' model permits the separation of trees that
cannot be separated by simpler methods, such as maximum likelihood or
classical active contours.

In the rest of this section, we present a brief introduction to HOACs. In
section~\ref{sec.circdet}, we describe the `gas of circles' HOAC model. The
key to this model is the analysis of the stability of a circle as a
function of the model parameters. To demonstrate the prior knowledge
contained in the model, and the empirical correctness of the stability
analysis, we present experimental results using the new energy. In
section~\ref{sec.expresult}, we apply the new model to tree crown
extraction. We describe a likelihood energy for trees based on the image
intensity and gradient, and then present experimental results on synthetic
data and on aerial images. We conclude in section~\ref{sec:conclusion}, and
discuss some open issues with the model.

\subsection{Higher order active contours}
\label{sec.highordintro}

As described in section~\ref{sec.introduction}, HOACs introduce long-range
interactions between boundary points not via the intermediary of a template
region or regions to which $\reg$ is compared, but directly, by using
energy terms that involve multiple integrals over the boundary. The
integrands of such integrals thus depend on two or more, perhaps widely
separated, boundary points simultaneously, and can thereby impose relations
between tuples of points. Euclidean invariance of such energies can be
imposed directly on the energy, without the necessity to estimate a
transformation between the boundary sought and the template. Perhaps more
importantly, because there is no template, the topology of the region need
not be constrained, a factor that is critical when the topology is not
known {\em a priori}.

As with all active contour models, a region $\reg$ is represented by its
boundary, $\bound$. There are various ways to think of the boundary of a
region. If the region has only one connected component, which is also
simply-connected, then a boundary is an equivalence class of embeddings of
the circle $\sircle$ under the action of orientation-preserving
diffeomorphisms of $\sircle$. When more, possibly multiply-connected
components are included, however, things get complicated. First, the number
of embeddings of $\sircle$ that are required depends on the topology, and
second, there are constraints on the orientations of different components
if they are to represent regions with handles.

An alternative is to view $\bound$ as a closed $1$-chain $\contour$ in the
image domain $\imdom$ (\cite{Choquet-Bruhat96} is a useful reference for
the following discussion). Although region boundaries correspond to a
special subset of closed $1$-chains known as domains of integration, active
contour energies themselves are defined for general $1$-chains. It is
convenient to use this more general context to distinguish HOAC energies
from classical active contours, because it allows for notions of linearity
to be used to characterize the complexity of energy functionals.

Using this representation, HOAC energies can be defined as follows
\citep{Rochery05d,Rochery06}. Let $\contour$ be a $1$-chain in $\imdom$,
and $\dom\contour$ be its domain. Then $\contour^{n}:
(\dom\contour)^{n}\map\imdom^{n}$ is an $n$-chain in $\imdom^{n}$. We
define a class of $(n - p)$-forms on $\imdom^{n}$ that are $1$-forms with
respect to $(n- p)$ factors and $0$-forms with respect to the remaining $p$
factors (by symmetry, it does not matter which $p$ factors). These forms
can be pulled back to $(\dom\contour)^{n}$ by $\contour^{n}$. The Hodge
duals of the $p$ $0$-form factors with respect to the induced metric on
$\dom\contour$ can then be taken independently on each such factor, thus
converting them to $1$-forms, and rendering the whole form an $n$-form on
$(\dom\contour)^{n}$. This $n$-form can then be integrated on
$(\dom\contour)^{n}$.

In the $(n, p) = (n, 0)$ cases, we are simply integrating a general
$n$-form on the image of $\contour^{n}$ in $\imdom^{n}$, thus defining a
linear functional on the space of $n$-chains in $\imdom^{n}$, and hence an
$n^{\text{th}}$-order monomial on the space of $1$-chains in $\imdom$.
Taking arbitrary linear combinations of such monomials then gives the space
of polynomial functionals on the space of $1$-chains. By analogy we refer
to the general $(n, p)$ cases as `generalized $n^{\text{th}}$-order
monomials' on the space of $1$-chains in $\imdom$, and to arbitrary linear
combinations of the latter as `generalized polynomial functionals' on the
space of $1$-chains in $\imdom$. HOAC energies are generalized polynomial
functionals. Standard active contour energies are generalized {\em linear}
functionals on $1$-chains in this sense, hence the term `higher-order'.

The $(1, 1)$ case is simply the boundary length in some metric. An
interesting application of the $(2, 2)$ case to topology preservation is
described by~\citet{Sundaramoorthi05}. The $(1, 0)$ case gives the region
area in some metric.

To be more concrete, we specialize to the $(n, 0)$ case. Let $F$ be an
$n$-form on $\imdom^{n}$. We pull $F$ back to the domain of $\contour^{n}$
and integrate it:
    \begin{equation}\label{eq.high.func}
        E(\contour) = \int_{(\partial R)^{n}} F
        = \int_{(\dom \contour)^{n}}(\contour^{n})^{\ast}
        F \eqstop
    \end{equation}

\noi Specializing again to the case $n = 2$, and using the antisymmetry of
$F$ together with the symmetry of $\contour^{2}$, we can rewrite the energy
functional in this case as
    \begin{equation}\label{eg.quad.func}
        E(\contour) = \int_{(\partial\reg)^{2}}F
        = \int_{(\dom \contour)^{2}}(\contour \times \contour)^{*} F
            = \iint_{(\dom \contour)^{2}} dt\intspace dt' \intspace
            \tanvec(t) \cdot F(\contour(t), \contour(t')) \cdot \tanvec(t')
            \eqcomma
    \end{equation}

\noi where $F(x, x')$, for each $(x, x') \in \imdom^{2}$, is a $2\times 2$
matrix, $t$ is a coordinate on $\dom\contour$, and $\tanvec =
\dot{\contour}$ is the tangent vector to $\contour$.

By imposing Euclidean invariance on this term, and adding linear terms,
\citet{Rochery03b} defined the following higher-order active contour prior:
    \begin{equation} \label{eq.q.1}
        \Eg(\contour)=\lambdac \length(\contour) + \alphac \area(\contour)
        - \frac{\betac}{2} \iint dt\intspace dt'\intspace
        \tanvec(t') \cdot \tanvec(t)\intspace \interactionfunction(R(t, t'))
        \eqcomma
    \end{equation}

\noi where $\length$ is the boundary length functional, $\area$ is the
interior area functional and $R(t, t')=\abs{\contour(t) - \contour(t')}$ is
the Euclidean distance between $\contour(t)$ and $\contour(t')$.
\citet{Rochery03b} used the following interaction function
$\interactionfunction$:
    \begin{equation}
        \interactionfunction(z) =
        \begin{cases}
            1 & \text{$z < \dmin - \epsilon$} \eqcomma\\
            \frac{1}{2}
            \bigl(
                1 - \frac{z - \dmin}{\epsilon} - \frac{1}{\pi}\sin \frac{\pi (z - \dmin)}{\epsilon}
            \bigr)
            & \text{$\dmin - \epsilon \leq z < \dmin + \epsilon$} \eqcomma \\
            0 & \text{$z \geq \dmin + \epsilon$} \eqstop
        \end{cases}
        \label{eq:interactionfunction}
    \end{equation}

In this paper, we use this same interaction function with $\dmin =
\epsilon$, but other monotonically decreasing functions lead to
qualitatively similar results.

\section{The `gas of circles' model} \label{sec.circdet}

For certain ranges of the parameters involved, the
energy~\eqref{eq.q.1} favours regions in the form of networks,
consisting of long narrow arms with approximately parallel sides,
joined together at junctions, as described by
\citet{Rochery03b,Rochery05d,Rochery06}. It thus provides a good
prior for network extraction from images. This behaviour does not
persist for all parameter values, however, and we will exploit this
parameter dependence to create a model for a `gas of circles', an
energy that favours regions composed of an {\em a priori} unknown
number of circles of a certain radius.

For this to work, a circle of the given radius, hereafter denoted $\ro$,
must be stable, that is, it must be a local minimum of the energy. In
section~\ref{sec.stab_anal}, we conduct a stability analysis of a circle,
and discover that stable circles are indeed possible provided certain
constraints are placed on the parameters. More specifically, we expand the
energy $\Eg$ in a functional Taylor series to second order around a circle
of radius $\ro$. The constraint that the circle be an energy extremum then
requires that the first order term be zero, while the constraint that it be
a minimum requires that the operator in the second order term be positive
semi-definite. These requirements constrain the parameter values. In
subsection~\ref{sec.geomexp}, we present numerical experiments using $\Eg$
that confirm the results of this analysis.

\subsection{Stability analysis} \label{sec.stab_anal}

We want to expand the energy $\Eg$ around a circle of radius $\ro$. We
denote a member of the equivalence class of maps representing the $1$-chain
defining the circle by $\contouro$. The energy $\Eg$ is invariant to
diffeomorphisms of $\dom\contouro$, and thus is well-defined on $1$-chains.
To second order,
    \begin{equation}
        \Eg (\contour) = \Eg(\contouro + \dcontour) = \Eg(\contouro)
        + \braket{\dcontour}{\frac{\delta \Eg}{\dcontour}}_{\contouro}
        + \frac{1}{2}\braoket{\dcontour}{\frac{\delta^{2} \Eg}{\dcontour^{2}}}{\dcontour}_{\contouro}
        \eqstop \label{eq:genericexpansion}
    \end{equation}

\noi where $\braket{\cdot}{\cdot}$ is a metric on the space of $1$-chains.

Since $\contouro$ represents a circle, it is easiest to express it in terms
of polar coordinates $r, \theta$ on $\imdom$. For a suitable choice of
coordinate on $\sircle$, a circle of radius $\ro$ centred on the origin is
then given by $\contouro(t) = (\ro(t), \thetao(t))$, where $\ro(t) =  \ro$,
$\theta(t) = t$, and $t \in [-\pi, \pi)$. We are interested in the
behaviour of small perturbations $\dcontour = (\dr, \delta\theta)$. The
first thing to notice is that because the energy $\Eg$ is defined on
$1$-chains, tangential changes in $\contour$ do not affect its value. We
can therefore set $\delta\theta=0$, and concentrate on $\dr$.

On the circle, using the arc length parameterization $t$, the
integrands of the different terms in $\Eg$ are functions of $t - t'$
only; they are invariant to translations around the circle. In
consequence, the second derivative
$\delta^{2}\Eg/\dcontour(t)\dcontour(t')$ is also translation
invariant, and this implies that it can be diagonalized in the
Fourier basis of the tangent space at $\contouro$. It thus turns out
to be easiest to perform the calculation by expressing $\dr$ in terms
of this basis:
    \begin{equation}
        \dr(t) = \sum_{k} a_{k} e^{i\ro kt}
        \eqcomma \label{eq:radius}
    \end{equation}

\noi where $k \in \{m /\ro \st m \in \integers\}$. Below, we simply state
the resulting expansions to second order in the $a_{k}$ for the three terms
appearing in equation~\eqref{eq.q.1}. Details can be found in
appendix~\ref{AppendixA}.

The boundary length and interior area of the region are given to second
order by
    \begin{align}
        \length(\contour) & = \int_{-\pi}^{\pi} dt \intspace \abs{\tanvec(t)}
        = 2\pi \ro  \left\{1+\frac{a_{0}}{\ro}
        + \frac{1}{2}\sum_{k} k^{2} \abs{a_{k}}^{2} \right\} \label{eq.f.l} \\
        \area(\contour) & = \int_{-\pi}^{\pi} d\theta \int_{0}^{r(\theta)} dr'\intspace r'
        = \pi \ro ^{2} + 2\pi \ro  a_{0}
        + \pi \sum_{k} \abs{a_{k}}^{2} \eqstop\label{eq.f.a}
    \end{align}

\noi Note the $k^{2}$ in the second order term for $L$. This is the same
frequency dependence as the Laplacian, and shows that the length term plays
a similar smoothing role for boundary perturbations as the Laplacian does
for functions. In the area term, by contrast, the Fourier perturbations are
`white noise'.

It is also worth noting that there are no stable solutions using these
terms alone. For the circle to be an extremum, we require $\lambdac 2\pi +
\alphac 2\pi \ro = 0$, which tells us that $\alphac = -\lambdac/\ro$. The
criterion for a minimum is, for each $k$, $\lambdac \ro k^{2} + \alphac
\geq 0$. Note that we must have $\lambdac
> 0$ for stability at high frequencies. Substituting for $\alphac$,
the condition becomes $\lambdac(\ro k^{2} - \ro^{-1}) \geq 0$. Substituting
$k = m/\ro$, gives the condition $m^{2} - 1 \geq 0$. Two points are worth
noting. The first is the one we have already made: the zero frequency
perturbation is not stable. The second is that the $m = 1$ perturbation is
marginally stable to second order, that is, such changes require no energy
to this order. To fully analyse them, we must therefore go to higher order
in the Taylor series. This feature will appear also in the analysis of the
full energy $\Eg$.

The quadratic term can be expressed to second order as
    \begin{multline}
        \iint_{-\pi}^{\pi} dt\intspace dt'\intspace
        \tanvec(t') \cdot \tanvec(t)\intspace \interactionfunction(R(t, t')) =
        2\pi \int_{-\pi}^{\pi} dp \intspace F_{00}(p)
        + 4\pi a_{0} \int_{-\pi}^{\pi} dp\intspace F_{10}(p) \\
        +\sum_{k} 2\pi \abs{a_{k}}^{2}
        \biggl\{
            \Bigl[
                2 \int_{-\pi}^{\pi} dp\intspace F_{20}(p)
                + \int_{-\pi}^{\pi} dp\intspace e^{-i\ro k p} F_{21}(p)
            \Bigr] \\
            -
            \Bigl[
                2 i\ro k \int_{-\pi}^{\pi} dp\intspace e^{-i\ro kp} F_{23}(p)
            \Bigr]
            +
            \Bigl[
                \ro ^{2}k^{2}\int_{-\pi}^{\pi} dp \intspace e^{-i \ro k p} F_{24}(p)
            \Bigr]
        \biggr\}
        \eqcomma \label{eq.f.q}
    \end{multline}

\noi The $F_{ij}$ are functionals of $\interactionfunction$ (hence
functions of $\dmin$ and $\epsilon$ for $\interactionfunction$ given by
equation~\eqref{eq:interactionfunction}), and functions of $\ro$, as well
as functions of the dummy variable $p$.

Combining equations~\eqref{eq.f.l},~\eqref{eq.f.a}, and~\eqref{eq.f.q}, we
find the energy functional~\eqref{eq.q.1} up to the second order:
    \begin{equation}
    \begin{split}
        \Eg(\contouro + \delta \contour)
        & = e_{0}(\ro) + a_{0}e_{1}(\ro) + \frac{1}{2}\sum_{k}\abs{a_{k}}^{2}
        e_{2}(k, \ro) \\
        & =
        \Bigl\{
            2\pi \lambdac \ro  + \pi \alphac \ro ^{2}
            - \pi\betac G_{00}(\ro)
        \Bigr\}
        + a_{0}
        \Bigl\{
            2\pi\lambdac  + 2\pi \alphac \ro - 2\pi\betac G_{10}(\ro)
        \Bigr\} \\
        & + \frac{1}{2}\sum_{k} \abs{a_{k}}^{2}
        \Bigl\{
            2\pi \lambdac \ro  k^{2} + 2\pi\alphac \\
            & - 2\pi\betac
            \bigl[
                2 G_{20}(\ro) + G_{21}(k, \ro)
                - 2 i \ro k G_{23}(k, \ro) + \ro ^{2}k^{2} G_{24}(k, \ro)
            \bigr]
        \Bigr\} \eqcomma
    \end{split}\label{eq:fullenergytosecondorder}
    \end{equation}

\noi where $G_{ij} = \int_{-\pi}^{\pi}dp\intspace e^{-i\ro (1 -
\delta(j)) k p} F_{ij}(p)$. Note that as anticipated, there are no
off-diagonal terms linking $a_{k}$ and $a_{k'}$ for $k \neq k'$: the
Fourier basis diagonalizes the second order term.

\subsubsection{Parameter constraints}
\label{sec:parameterconstraints}

Note that a circle of any radius is always an extremum for non-zero
frequency perturbations ($a_{k}$ for $k \neq 0$), as these Fourier
coefficients do not appear in the first order term (this is also a
consequence of invariance to translations around the circle). The condition
that a circle be an extremum for $a_{0}$ as well ($e_{1} = 0$) gives rise
to a relation between the parameters:
    \begin{equation}
        \betac(\lambdac, \alphac, \hatro ) =
        \frac{\lambdac + \alphac \hatro}{G_{10}(\hatro)}
        \eqcomma\label{eq.beta}
    \end{equation}

\begin{figure}[!h]
    \begin{center}
        \begin{tabular}{c}
            \includegraphics[width=0.45\textwidth]{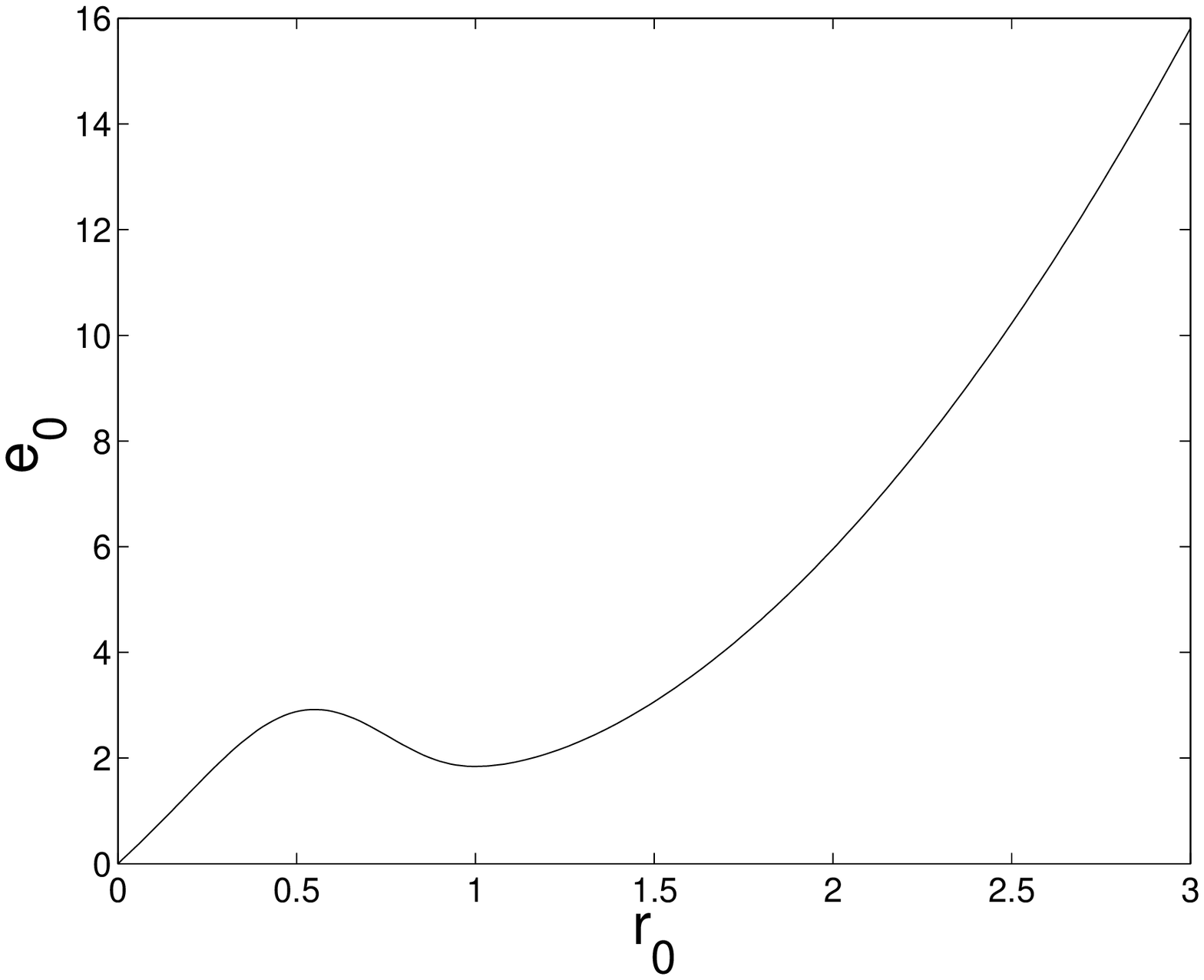}
        \end{tabular}
        \begin{tabular}{c}
            \includegraphics[width=0.45\textwidth]{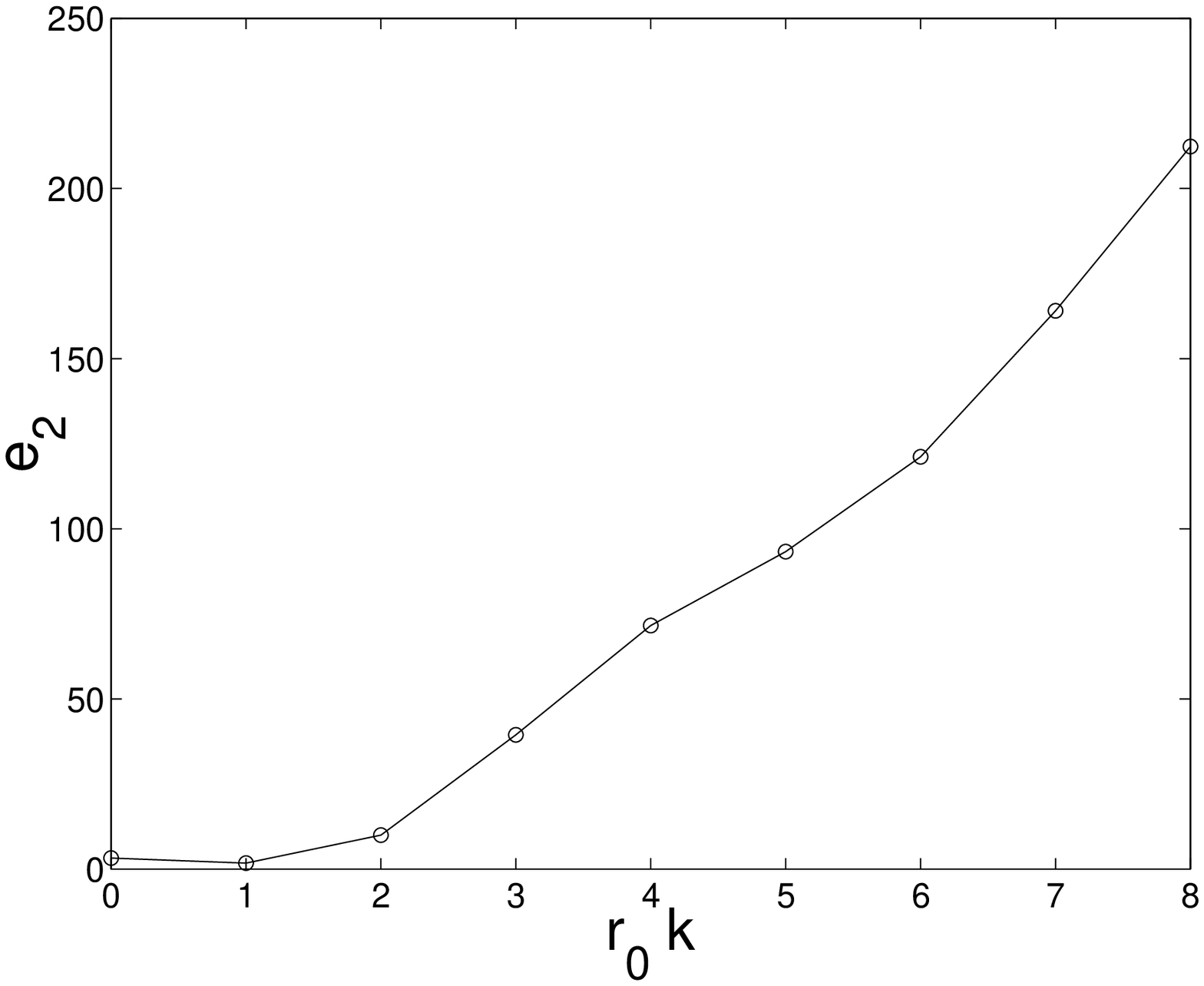}
        \end{tabular}
    \end{center}
    \caption{Plots of $e_{0}$ against $\ro$ and $e_{2}$ against $\hatro k$.
    Left: the energy of a circle $e_{0}$ plotted against radius $\ro$ for
    $\lambdac = 1.0$, $\alpha=0.8$, and $\betac=1.39$ calculated from
    equation~\eqref{eq.beta} with $\hatro = 1.0$. (The parameters of
    $\interactionfunction$ are $\dmin = 1.0$ and $\epsilon = 1.0$, but note
    that it is not necessary in general that $\dmin = \hatro$.) The
    function has a minimum at $\ro = \hatro$ as desired. Right: the second
    derivative of $\Eg$, $e_{2}$, plotted against $\hatro k$ for the same
    parameter values. The function is non-negative for all
    frequencies.}\label{fig:e0e2}
\end{figure}

\begin{figure}[!h]
    \begin{center}
        \begin{tabular}{c}
            \includegraphics[width=0.5\textwidth]{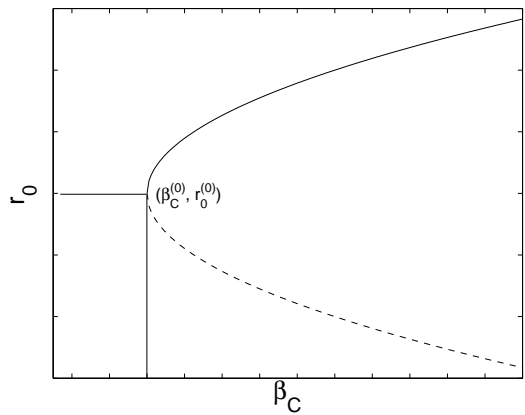}
        \end{tabular}
    \end{center}
    \caption{Schematic plot of the positions of the extrema of the energy
    of a circle versus $\betac$.}
    \label{fig:catastrophecurve}
\end{figure}

\noi where we have introduced $\hatro$ to indicate the radius at which
there is an extremum, to distinguish it from $\ro$, the radius of the
circle about which we are calculating the
expansion~\eqref{eq:genericexpansion}. The left hand side of
figure~\ref{fig:e0e2} shows a typical plot of the energy $e_{0}$ of a
circle versus its radius $\ro$, with the $\betac$ parameter fixed using the
equation~\eqref{eq.beta} with $\lambdac = 1.0$, $\alpha=0.8$, and $\hatro =
1.0$. The energy has a minimum at $\ro = \hatro$ as desired. The
relationship between $\hatro$ and $\betac$ is not quite as straightforward
as it might seem though. As can be seen, the energy also has a maximum at
some radius. It is not {\em a priori} clear whether it will be the maximum
or the minimum that appears at $\hatro$. If we graph the positions of the
extrema of the energy of a circle against $\betac$ for fixed $\alphac$, we
find a curve qualitatively similar to that shown in
figure~\ref{fig:catastrophecurve} (this is an example of a fold
catastrophe). The solid curve represents the minimum, the dashed the
maximum. Note that there is indeed a unique $\betac$ for a given choice of
$\hatro$. Denote the point at the bottom of the curve by $(\betac^{(0)},
\hatro^{(0)})$. Note that at $\betac = \betac^{(0)}$, the extrema merge and
for $\betac < \betac^{(0)}$, there are no extrema: the energy curve is
monotonic because the quadratic term is not strong enough to overcome the
shrinking effect of the length and area terms. Note also that the minimum
cannot move below $\ro = \ro^{(0)}$. This behaviour is easily understood
qualitatively in terms of the interaction function in
equation~\eqref{eq:interactionfunction}. If $2\ro < \dmin - \epsilon$, the
quadratic term will be constant, and no force will exist to stabilize the
circle. In order to use equation~\eqref{eq.beta} then, we have to ensure
that we are on the upper branch of figure~\ref{fig:catastrophecurve}.

Equation~\eqref{eq.beta} gives the value of $\betac$ that provides an
extremum of $e_{0}$ with respect to changes of radius $a_{0}$ at a given
$\hatro$ ($e_{1}(\hatro) = 0$), but we still need to check that the circle
of radius $\hatro$ is indeed stable to perturbations with non-zero
frequency, \ie that $e_{2}(k, \hatro)$ is non-negative for all $k$. Scaling
arguments mean that in fact the sign of $e_{2}$ depends only on the
combinations $\tro = \ro/\dmin$ and $\talphac = (\dmin/\lambdac)\alphac$.
The equation for $e_{2}$ can then be used to obtain bounds on $\talphac$ in
terms of $\tro$. (Details of these calculations and bounds will be given
elsewhere.) The right hand side of figure~\ref{fig:e0e2} shows a plot of
$e_{2}(k, \hatro)$ against $\hatro k$ for the same parameter values used
for the right hand side, showing that it is non-negative for all $\hatro
k$.

We call the resulting model, the energy $\Eg$ with parameters chosen
according to the above criteria, the `gas of circles' model.

\subsection{Geometric experiments}
\label{sec.geomexp}

\begin{figure}[!h]
    \begin{center}
        \begin{tabular}{|c||c|c|c|}
        \hline
            \includegraphics[width=0.15\hsize]{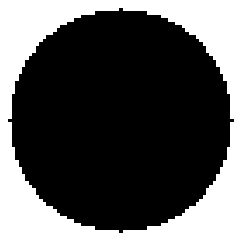} &
            \includegraphics[width=0.15\hsize]{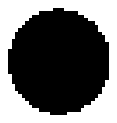} &
            \includegraphics[width=0.15\hsize]{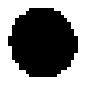} &
            \includegraphics[width=0.15\hsize]{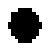} \\
        \hline
            \includegraphics[width=0.15\hsize]{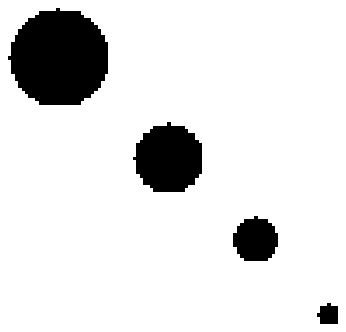} &
            \includegraphics[width=0.15\hsize]{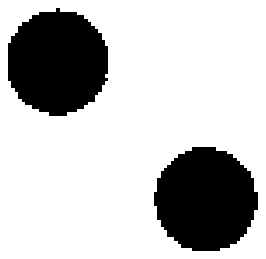} &
            \includegraphics[width=0.15\hsize]{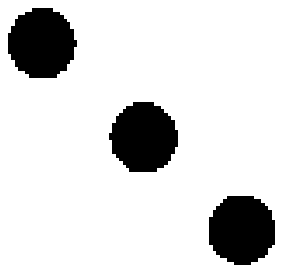} &
            \includegraphics[width=0.15\hsize]{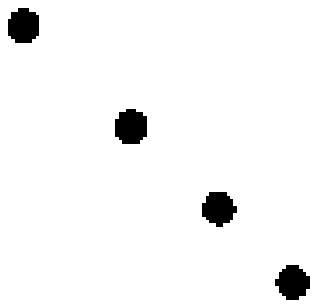} \\
        \hline
            \includegraphics[width=0.15\hsize]{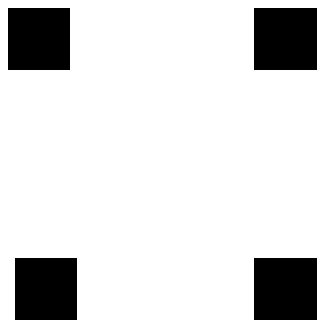} &
            \includegraphics[width=0.15\hsize]{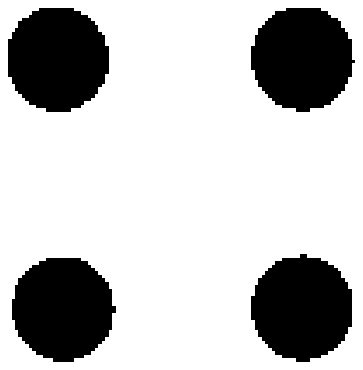} &
            \includegraphics[width=0.15\hsize]{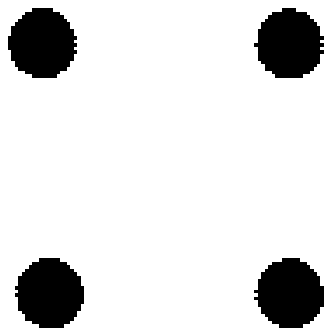} &
            \includegraphics[width=0.15\hsize]{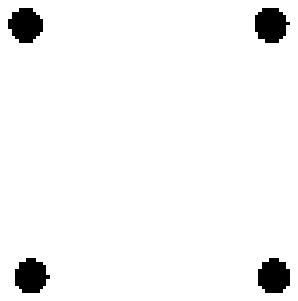} \\
        \hline
            \includegraphics[width=0.15\hsize]{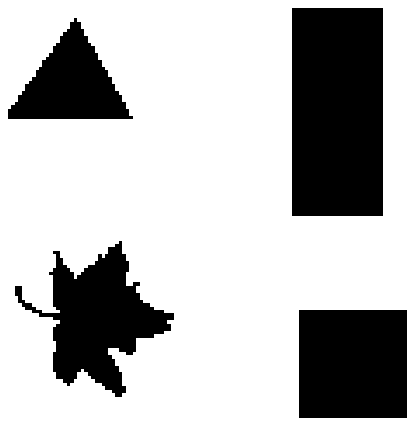} &
            \includegraphics[width=0.15\hsize]{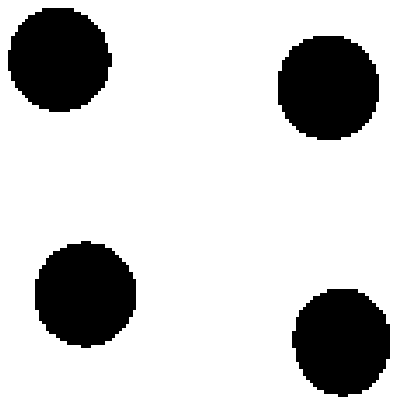} &
            \includegraphics[width=0.15\hsize]{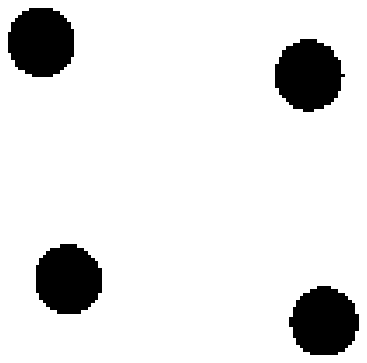} &
            \includegraphics[width=0.15\hsize]{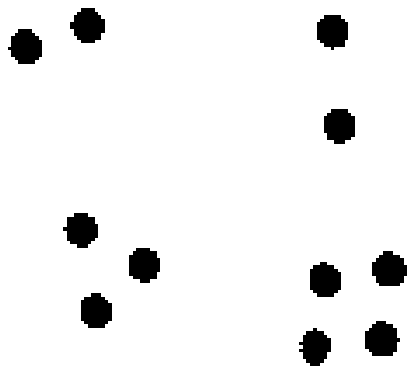} \\
        \hline
        \multicolumn{1}{c}{(Initial)} & \multicolumn{1}{c}{($\hatro =15$)}
        & \multicolumn{1}{c}{($\hatro =10$)} & \multicolumn{1}{c}{($\hatro =5$)}
        \end{tabular}
    \end{center}
    \caption{Experimental results using the geometric term: the first column
    shows the initial conditions; the other columns show the stable states for
    various choices of the radius.} \label{fig:geom}
\end{figure}

In order to illustrate the behaviour of the prior energy $\Eg$ with
parameter values fixed according to the above analysis, in this section we
show the results of some experiments using this energy (there are no image
terms). Figure~\ref{fig:geom} shows the result of gradient descent using
$\Eg$ starting from various different initial regions. (For details of the
implementation of gradient descent for higher-order active contour energies
using level set methods, see \citep{Rochery05d,Rochery06}.) In the first
column, four different initial regions are shown. The other three columns
show the final regions, at convergence, for three different sets of
parameters. In particular, the three columns have $\hatro = 15.0$, $10.0$,
and $5.0$ respectively.

In the first row, the initial shape is a circle of radius $32$ pixels. The
stable states, which can be seen in the other three columns, are circles
with the desired radii in every case. In the second row, the initial region
is composed of four circles of different radii. Depending on the value of
$\hatro$, some of these circles shrink and disappear. This behaviour can be
explained by looking at figure~\ref{fig:e0e2}. As already noted, the energy
of a circle $e_{0}$ has a maximum at some radius $r_{\text{max}}$. If an
initial circle has a radius less than $r_{\text{max}}$, it will `slide down
the energy slope' towards $\ro = 0$, and disappear. If its radius is larger
than $r_{\text{max}}$, it will finish in the minimum, with radius $\hatro$.
This is precisely what is observed in this second experiment. In the third
row, the initial condition is composed of four squares. The squares evolve
to circles of the appropriate radii. The fourth row has an initial
condition composed of four differing shapes. The nature of the stable
states depends on the relation between the stable radius, $\hatro$, and the
size of the initial shapes. If $\hatro$ is much smaller than an initial
shape, this shape will `decay' into several circles of radius $\hatro$.

\section{Data terms and experiments} \label{sec.expresult}

In  this section, we apply the `gas of circles' model developed in
section~\ref{sec.circdet} to the extraction of trees from aerial images.
This is just one of many possible applications, corresponding to the
mission of the Ariana research group. In the next section, we give a brief
state of the art for tree crown extraction, and then present the data terms
we use in section~\ref{sec.dataterm}. In
section~\ref{sec:treecrownexperiments}, we describe tree crown extraction
experiments on aerial images and compare the results to those found using a
classical active contour model. In section~\ref{sec:noiseexperiments}, we
examine the robustness of the method to noise using synthetic images. This
illuminates the principal failure modes of the model, which will be further
discussed in section~\ref{sec:conclusion}, and which point the way for
future work. In section~\ref{sec:circleseparation}, we illustrate the
importance of prior information via tree crown separation experiments on
synthetic images, and compare the results to those obtained using a
classical active contour model.

\subsection{Previous work}
\label{sec:stateofthearttreecrowns}

The problem of locating, counting, or delineating individual trees in high
resolution aerial images has been studied in a number of papers.

\citet{Gougeon95a,Gougeon98a} observes that trees are brighter than the
areas separating them. Local minima of the image are found using a $3\times
3$ filter, and the `valleys' connecting them are then found using a
$5\times 5$ filter. The tree crowns are subsequently delineated using a
five-level rule-based method designed to find circular shapes, but with
some small variations permitted. \citet{Larsen98,Larsen99} concentrates on
spruce tree detection using a template matching method. The main difference
between these two papers is the use of multiple templates in the second.
The 3D shape of the tree is modelled using a generalized ellipsoid, while
illumination is modelled using the position of the sun and a clear-sky
model. Reflectance is modelled using single reflections, with the branches
and needles acting as scatterers, while the ground is treated as a
Lambertian surface. Template matching is used to calculate a correlation
measure between the tree image predicted by the model and the image data.
The local maxima of this measure are treated as tree candidates, and
various strategies are then used to eliminate false positives.
\citet{Brandtberg98} decompose an image into multiple scales, and then
define tree crown boundary candidates at each scale as zero crossings with
convex greyscale curvature. Edge segment centres of curvature are then used
to construct a candidate tree crown region at each scale. These are then
combined over different scales and a final tree crown region is grown.
\citet{Andersen01} use a morphological approach combined with a top-hat
transformation for the segmentation of individual trees.

All of these methods use multiple steps rather than a unified model. Closer
in spirit to the present work is that of \citet{Perrin04,Perrin05}, who
model the collection of tree crowns by a marked point process, where the
marks are circles or ellipses. An energy is defined that penalizes, for
example, overlapping shapes, and controls the parameters of the individual
shapes. Reversible Jump MCMC and simulated annealing are used to estimate
the tree crown configuration. Compared to the work described in this paper,
the method has the advantage that overlapping trees can be represented as
two separate objects, but the disadvantage that the tree crowns are not
precisely delineated due to the small number of degrees of freedom for each
mark.

\subsection{Data terms and gradient descent} \label{sec.dataterm}

In order to couple the region model $\Eg$ to image data, we need a
likelihood, $\pr{\imag | \reg, \knowledge}$. The images we use for
the experiments are coloured infrared (CIR) images. Originally they
are composed of three bands, corresponding roughly to green, red, and
near infrared (NIR). Analysis of the one-point statistics of the
image in the region corresponding to trees and the image in the
background, shows that the `colour' information does not add a great
deal of discriminating power compared to a `greyscale' combination of
the three bands, or indeed the NIR band on its own. We therefore
model the latter.

The images have a resolution $\sim 0.5$m/pixel, and tree crowns have
diameters of the order of ten pixels. Very little if any dependence remains
between the pixels at this resolution, which means, when combined with the
paucity of statistics within each tree crown, that pixel dependencies (\ie
texture) are very hard to use for modelling purposes. We therefore model
the interior of tree crowns using a Gaussian distribution with mean $\muin$
and covariance $\sigmain^{2}\delta_{\reg}$, where $\delta_{A}$ is the
identity operator on images on $A\subset\imdom$.

The background is very varied, and thus hard to model in a precise way. We
use a Gaussian distribution with mean $\muout$ and variance
$\sigmaout^{2}\delta_{\regcmp}$. In general, $\muin > \muout$, and
$\sigmain < \sigmaout$; trees are brighter and more constant in intensity
than the background. The boundary of each tree crown has significant
inward-pointing image gradient, and although the Gaussian models should in
principle take care of this, we have found in practice that it is useful to
add a gradient term to the likelihood energy. Our likelihood thus has three
factors:
    \begin{equation}
        \pr{\imag | \reg, \knowledge} = Z^{-1}\intspace
        g_{\reg}(\imagin)\intspace g_{\regcmp}(\imagout)\intspace f_{\bound}(\imag_{\bound})
        \eqstop\non
    \end{equation}

\noi where $\imagin$ and $\imagout$ are the images restricted to
$\reg$ and $\regcmp$ respectively, and $g_{\reg}$ and $g_{\regcmp}$
are proportional to the Gaussian distributions already described, \ie
    \begin{equation}
        -\ln g_{\reg}(\imagin) = \int_{\reg} d^{2}x\intspace
        \frac{1}{2\sigmain^{2}}(\imagin(x)-\muin)^{2}
    \end{equation}

\noi and similarly for $g_{\regcmp}$. The function $f_{\bound}$ depends on
the gradient of the image $\del\imag$ on the boundary $\bound$:
    \begin{equation}
        -\ln f_{\bound}(\imag_{\bound}) =
        \lambdai \int_{\dom\contour} dt\intspace \normvec(t) \cdot \del I(t)
    \end{equation}

\noi where $\normvec$ is the unnormalized outward normal to
$\contour$. The normalization constant $Z$ is thus a function of
$\muin$, $\sigmain$, $\muout$, $\sigmaout$, and $\lambdai$. $Z$ is
also a functional of the region $\reg$. To a first approximation, it
is a linear combination of $\length(\bound)$ and $\area(\reg)$. It
thus has the effect of changing the parameters $\lambdac$ and
$\alphac$ in $\Eg$. However, since these parameters are essentially
fixed by hand (the criteria described in
section~\ref{sec:parameterconstraints} only allow us to fix $\betac$
and constrain $\alphac$), knowledge of the normalization constant
does not change their values, and we ignore it once the likelihood
parameters have been learnt.

The full model is then given by $E(\reg) = \Ei(\imag, \reg) + \Eg(\reg)$,
where
    \begin{equation}
        \Ei(\imag, \reg) = -\ln\pr{\imag | \reg, \knowledge} + \ln Z
        = -\ln g_{\reg}(\imagin) - \ln g_{\regcmp}(\imagout)
        -\ln f_{\bound}(\imag_{\bound})
        \eqstop\non
    \end{equation}

The energy is minimized by gradient descent. The gradient descent equation
for $E$ is
    \begin{multline}
        \normanormvec \cdot \frac{\partial \contour}{\partial \timeparam}(t) =
        - \del^{2} \imag(\contour(t)) - \frac{(\imag(\contour(t)) - \muin)^{2}}{2
        \sigmain^{2}} + \frac{(\imag(\contour(t)) - \muout)^{2}}{2
        \sigmaout^{2}} \\
        -\lambdac \curv(t) - \alphac + \betac \int_{\dom\contour} dt'\intspace \normaRvec(t, t') \cdot \normvec(t')
        \dot{\interactionfunction}(R(t, t')) \eqcomma \label{eq:quadraticforce}
    \end{multline}

\noi where $\timeparam$ is the descent time parameter, $\normaRvec(t, t') =
(\contour(t) - \contour(t'))/\abs{\contour(t) - \contour(t')}$ and $\curv$
is the signed boundary curvature. As already mentioned, to evolve the
region we use the level set framework of \citet{Osher88} extended to the
demands of nonlocal forces such as
equation~\eqref{eq:quadraticforce}~\citep{Rochery05d,Rochery06}.

\subsection{Tree crown extraction from aerial images}
\label{sec:treecrownexperiments}

In this section, we present the results of the application of the above
model to $50$ cm/pixel colour infrared aerial images of poplar stands
located in the `Sa{\^o}ne et Loire' region in France. The images were provided
by the French National Forest Inventory (IFN). As stated in
section~\ref{sec.dataterm}, we model only the NIR band of these images, as
adding the other two bands does not increase discriminating power. The tree
crowns in the images are $\sim 8$--$10$ pixels in diameter, \ie $\sim
4$--$5$m.

In the experiments, we compare our model to a classical active contour
model ($\betac = 0$). The parameters $\muin$, $\sigmain$, $\muout$, and
$\sigmaout$ were the same for both models, and were learned from
hand-labelled examples in advance. The classical active contour prior model
thus has three free parameters ($\lambdai$, $\lambdac$ and $\alphac$),
while the HOAC `gas of circles' model has six ($\lambdai$, $\lambdac$,
$\alphac$, $\betac$, $\dmin$ and $\ro$). We fixed $\ro$ based on our prior
knowledge of tree crown size in the images, and $\dmin$ was then set equal
to $\ro$. Once $\alphac$ and $\lambdac$ have been fixed, $\betac$ is
determined by equation~\eqref{eq.beta}. There are thus three effective
parameters for the HOAC model. In the absence of any method to learn
$\lambdai$, $\alphac$ and $\lambdac$, they were fixed by hand to give the
best results, as with most applications of active contour models. The
values of $\lambdai$, $\lambdac$ and $\alphac$ were not the same for the
classical active contour and HOAC models; they were chosen to give the best
possible result for each model separately.

The initial region in all real experiments was a rounded rectangle slightly
bigger than the image domain. The image values in the region exterior to
the image domain were set to $\muout$ to ensure that the region would
shrink inwards.

\begin{figure}[!h]
    \begin{center}
        \begin{tabular}{cc}
            \includegraphics[width=0.45\textwidth]{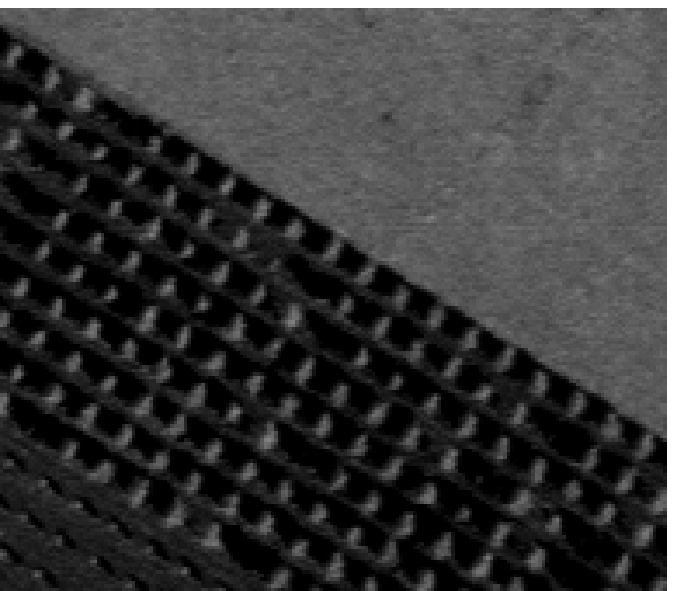} &
            \includegraphics[width=0.45\textwidth]{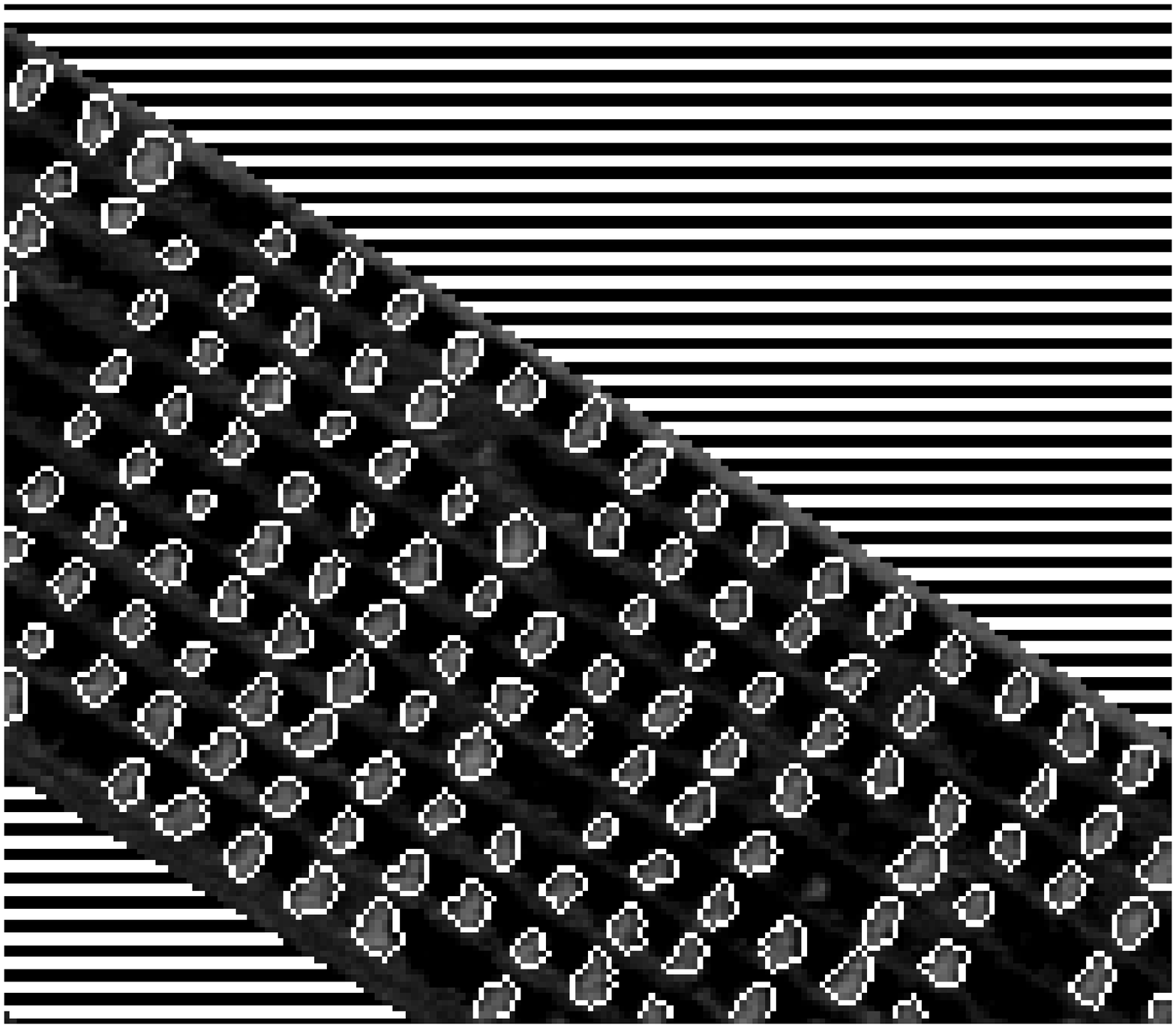}
        \end{tabular}
    \end{center}
    \caption{Left: real image with a planted forest \copyright IFN (0.3,
    0.06, 0.05, 0.05). Right: the result obtained using the `gas of
    circles' model $(529, 5.88, 5.88, 5.64, 4, 4)$.} \label{fig:real1}
\end{figure}

Figure~\ref{fig:real1} illustrates the first experiment. On the left is the
data, showing a regularly planted poplar stand. The result is shown on the
right. We have applied the algorithm only in the central part of the image,
for reasons that will be explained in section~\ref{sec:conclusion}.

\begin{figure}[!h]
    \begin{center}
        \begin{tabular}{ccc}
            \includegraphics[width=0.3\textwidth]{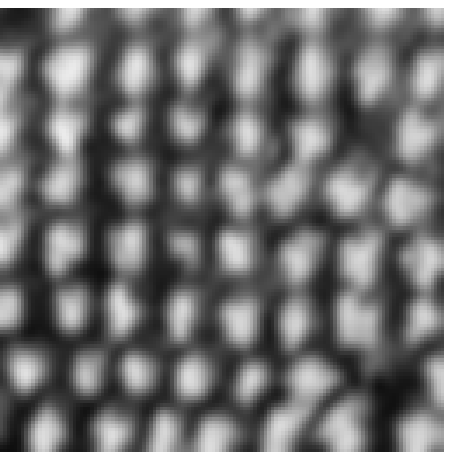} &
            \includegraphics[width=0.3\textwidth]{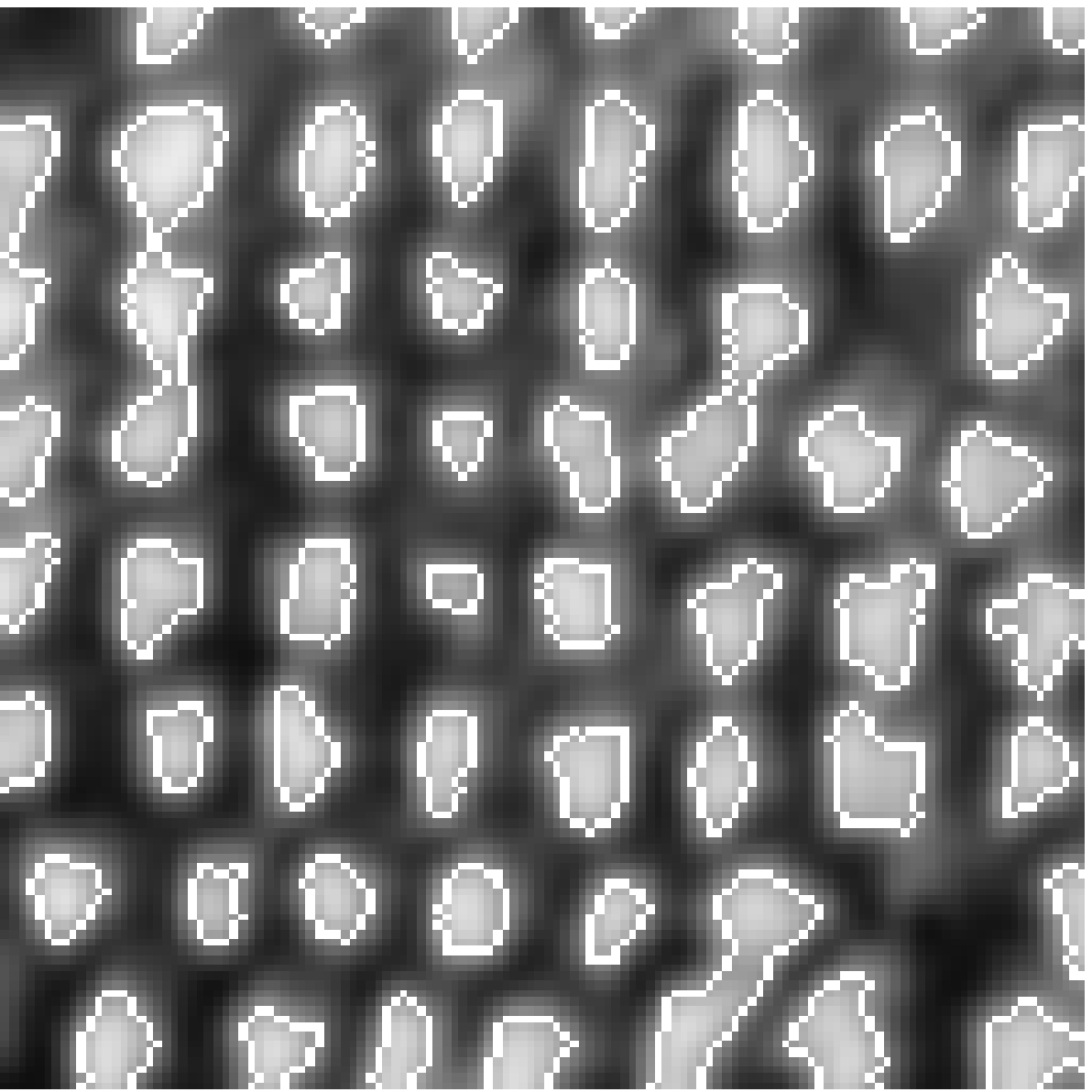} &
            \includegraphics[width=0.3\textwidth]{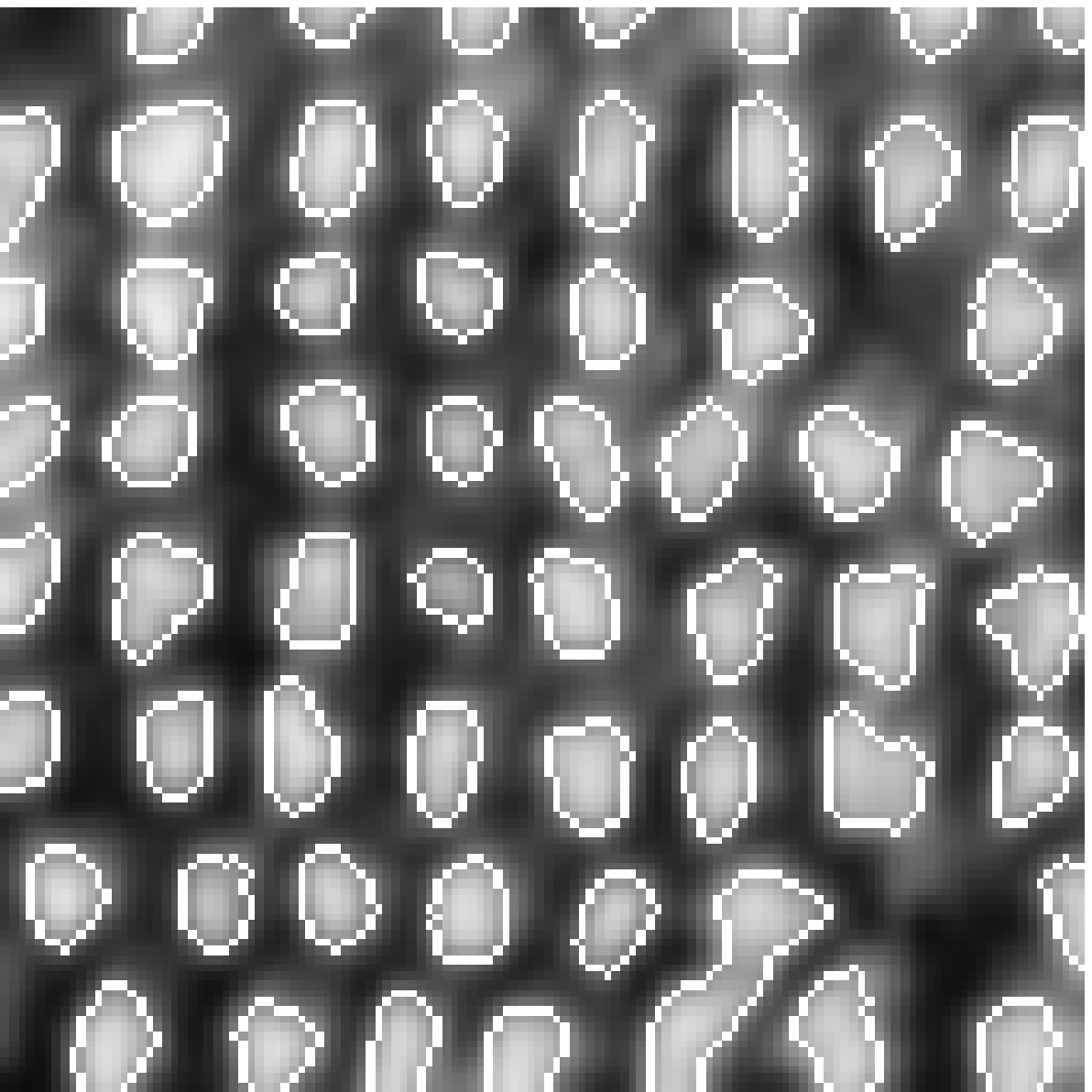}
        \end{tabular}
    \end{center}
    \caption{From left to right: image of poplars \copyright IFN (0.73,
    0.11, 0.23, 0.094); the best result with a classical active contour
    $(880, 13, 73)$; result with the 'gas of circles' model $(100, 6.7, 39,
    31, 4.2, 4.2)$.} \label{fig:expres1}
\end{figure}

Figure~\ref{fig:expres1} illustrates a second experiment. On the left is
the data. The image shows a small piece of an irregularly planted poplar
forest. The image is difficult because the intensities of the crowns are
varied and the gradients are blurred. In the middle is the best result we
could obtain using a classical active contour. On the right is the result
we obtain with the HOAC `gas of circles' model.\footnote{Unless otherwise
specified, in the figure captions the values of the parameters learned from
the image are shown when the data is mentioned, in the form $(\muin,
\sigmain, \muout, \sigmaout)$. The other parameter values are shown when
each result is mentioned, in the form $(\lambdai, \lambdac, \alphac,
\betac, \dmin, \ro)$, truncated if the parameters are not present. All
parameter values are truncated to two significant figures. Unless otherwise
specified, images were scaled to take values in $[0, 1]$. The region
boundary is shown in white.} Note that in the classical active contour
result several trees that are in reality separate are merged into single
connected components, and the shapes of trees are often rather distorted,
whereas the prior geometric knowledge included when $\beta \neq 0$ allows
the separation of almost all the trees and the regularization of their
shapes.

\begin{figure}[!h]
    \begin{center}
        \begin{tabular}{ccc}
            \includegraphics[width=0.3\textwidth]{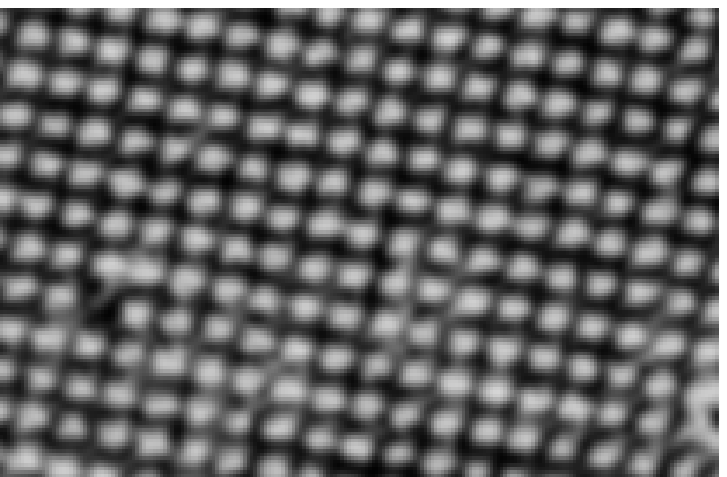} &
            \includegraphics[width=0.3\textwidth]{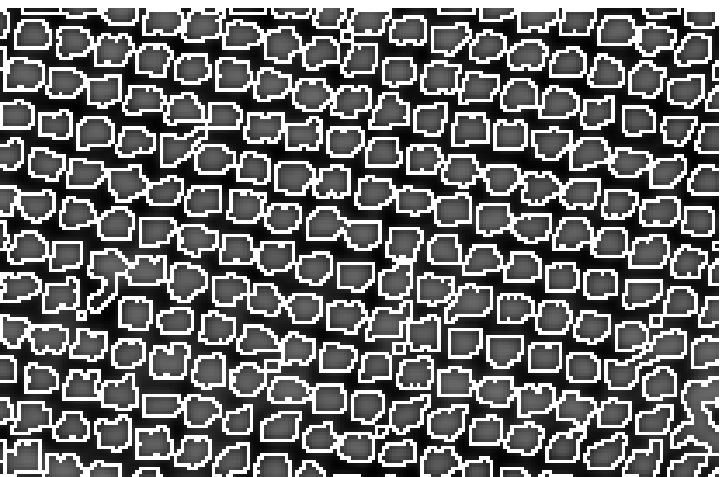} &
            \includegraphics[width=0.3\textwidth]{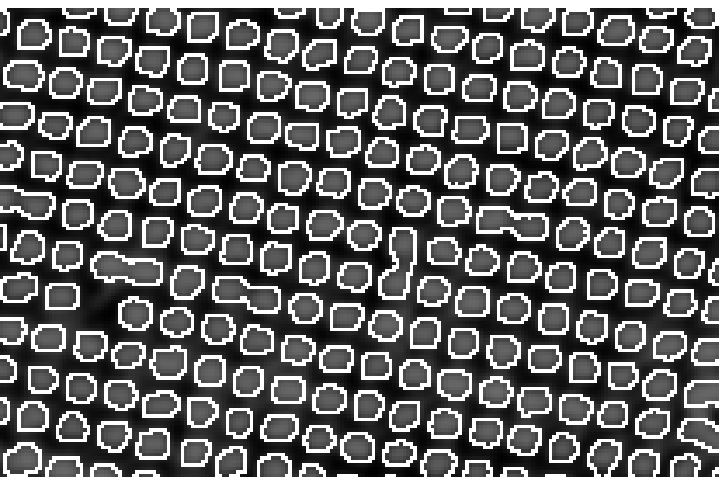}
        \end{tabular}
    \end{center}
    \caption{From left to right: image of poplars {\copyright}IFN (0.71, 0.075, 0.18,
    0.075); the best result with a classical active contour $(24000, 100,
    500)$; result with the 'gas of circles' model $(1500, 25, 130, 100,
    3.5, 3.5)$.} \label{fig:expres2}
\end{figure}

Figure~\ref{fig:expres2} illustrates a third experiment. Again the data is
on the left, the best result obtained with a classical active contour model
is in the middle, and the result with the HOAC `gas of circles' model is on
the right. The trees are closer together than in the previous experiment.
Using the classical active contour, the result is that the tree crown
boundaries touch in the majority of cases, despite their separation in the
image. Many of the connected components are malformed due to background
features. The HOAC model produces more clearly delineated tree crowns, but
there are still some joined trees. We will discuss this further in
section~\ref{sec:conclusion}

\begin{figure}[!h]
    \begin{center}
        \begin{tabular}{ccc}
            \includegraphics[width=0.18\textwidth]{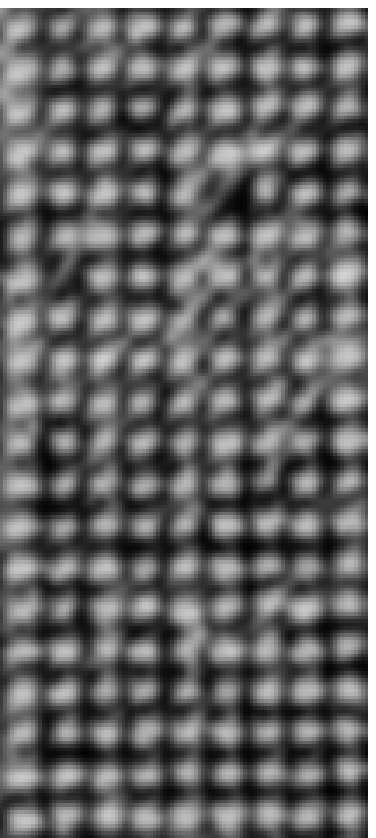} &
            \includegraphics[width=0.18\textwidth]{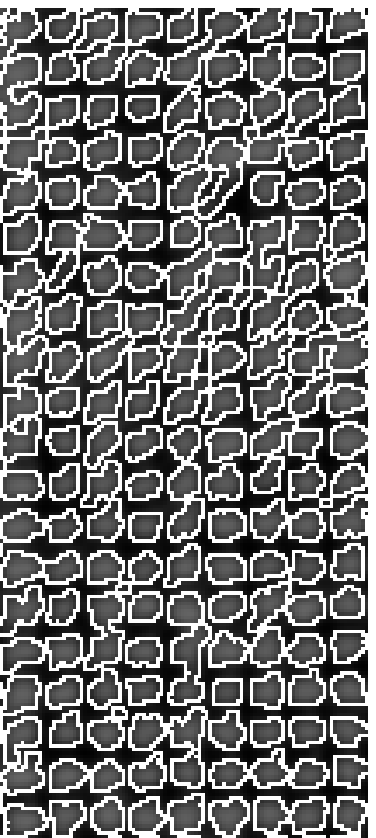} &
            \includegraphics[width=0.18\textwidth]{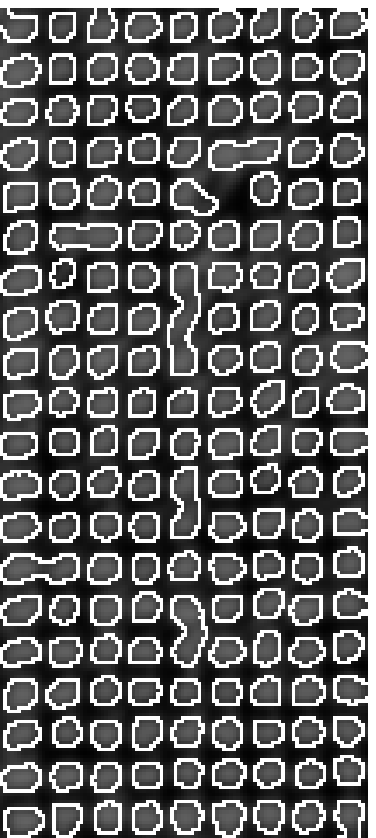}
        \end{tabular}
    \end{center}
    \caption{From left to right: image of poplars {\copyright}IFN (0.71, 0.075, 0.18,
    0.075); the best result with a classical active contour $(35000, 100,
    500)$; result with the 'gas of circles' model $(1200, 20, 100, 82, 3.5,
    3.5)$.} \label{fig:expres3}
\end{figure}

Figure~\ref{fig:expres3} shows a fourth experiment. The data is on the
left, the best result obtained with a classical active contour model is in
the middle, and the result with the HOAC `gas of circles' model is on the
right. Again, the `gas of circles' model better delineates the tree crowns
and separates more trees, but some joined trees remain also. The HOAC model
selects only objects of the size chosen, so that false positives involving
small objects do not occur.

\begin{table}
    \begin{center}
        \begin{tabular}{|c||c|c|c||c|c|c|}
            \hline
            Model &  & CAC &  &  & HOAC &  \\
            \hline
            Figure & CD \% & FP \% & FN \%  & CD \% & F+ \% & F- \% \\
            \hline
            Figure~\ref{fig:expres1} & 85 & 0 & 15  & 97 & 0 & 3 \\
            Figure~\ref{fig:expres2} & 96.2 & 2.8 & 1.9 & 97.7 & 0 & 2.3 \\
            Figure~\ref{fig:expres3} & 89.4 & 5 & 5.6 & 95.5 & 0.6 & 3.9 \\
            \hline
        \end{tabular}
    \end{center}
    \caption{Results on real images using a classical active contour model
    (CAC) and the `gas of circles' model (HOAC). CD: correct detections;
    FP: false positives; FN: false negatives (two joined trees count as one
    false negative).} \label{tab:real}
\end{table}

Table~\ref{tab:real} shows the percentages of correct tree detections,
false positives and false negatives (two joined trees count as one false
negative), obtained with the classical active contour model and the `gas of
circles' model in the experiments shown in
figures~\ref{fig:expres1},~\ref{fig:expres2}, and~\ref{fig:expres3}. The
`gas of circles' model outperforms the classical active contour in all
measures, except in the number of false negatives in the experiment in
figure~\ref{fig:expres2}.

Once the segmentation result has been obtained, it is a relatively simple
matter to compute statistics of interest to the forestry industry: number
of trees, total area, number and area density, and so on.

\subsection{Noisy synthetic images}
\label{sec:noiseexperiments}

In this section, we present the results of tests of the sensitivity
of the model to noise in the image. Fifty synthetic images were
created, each with ten circles with radius $8$ pixels and ten circles
with radius $3.5$ pixels, placed at random but with overlaps
rejected. Six different levels of white Gaussian noise, with image
variance to noise power ratios from $-5$ dB to $20$ dB, were then
added to the images to generate $300$ noisy images. Six of these,
corresponding to noisy versions of the same original image, were used
to learn $\muin$, $\sigmain$, $\muout$, and $\sigmaout$. The model
used was the same as that used for the aerial images, except that
$\lambdai$ was set equal to zero. The parameters were adjusted to
give a stable radius of $8$ pixels.
\begin{figure}[p]
    \begin{center}
        \begin{tabular}{cc}
            \includegraphics[width=0.32\textwidth]{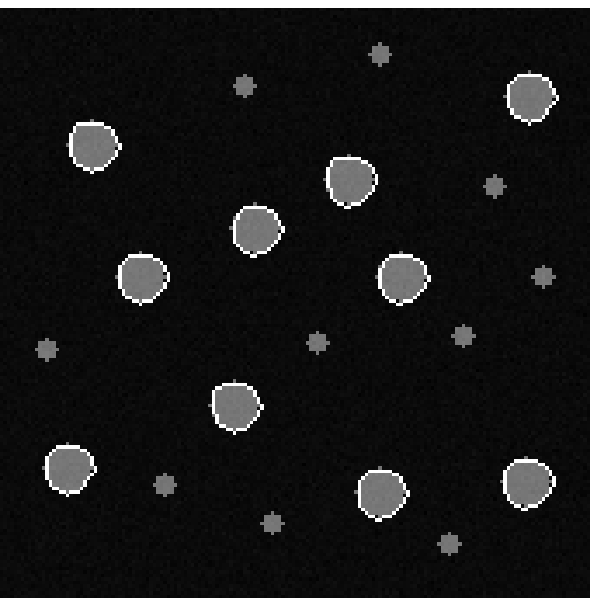} &
            \includegraphics[width=0.32\textwidth]{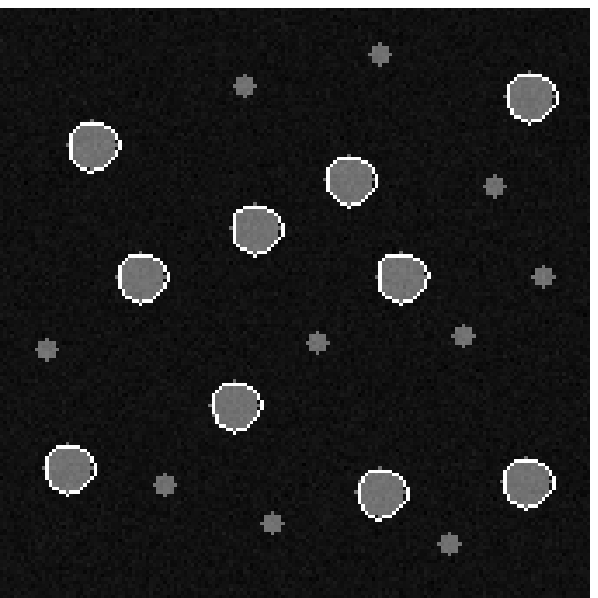} \\
            $(0.90, 0.028, 0.11, 0.028, 100, 100, 170)$ &
            $(0.85, 0.043, 0.16, 0.043, 33, 33, 58)$ \\
            \includegraphics[width=0.32\textwidth]{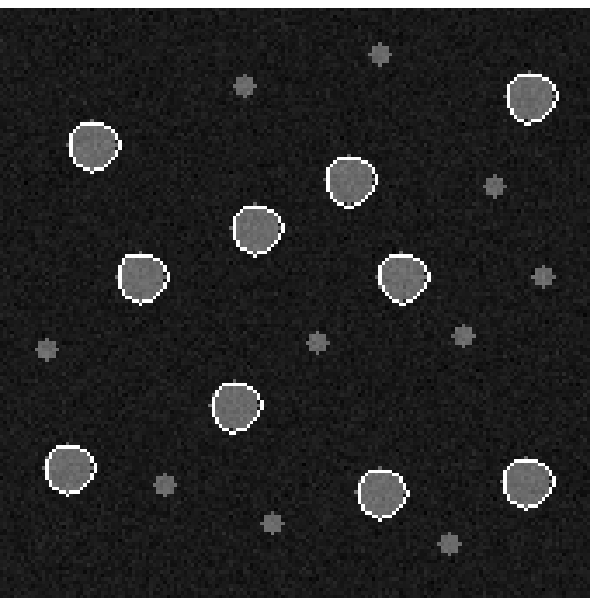} &
            \includegraphics[width=0.32\textwidth]{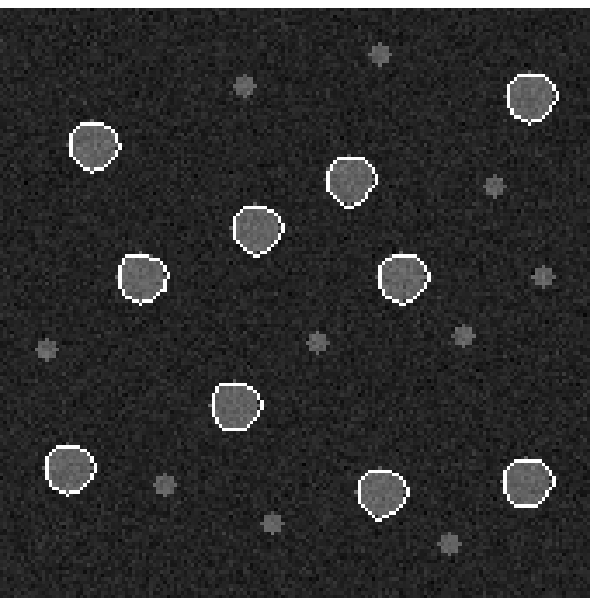} \\
            $(0.78, 0.061, 0.23, 0.062, 13, 13, 22)$ &
            $(0.71, 0.081, 0.31, 0.081, 4, 4, 6.9)$ \\
            \includegraphics[width=0.32\textwidth]{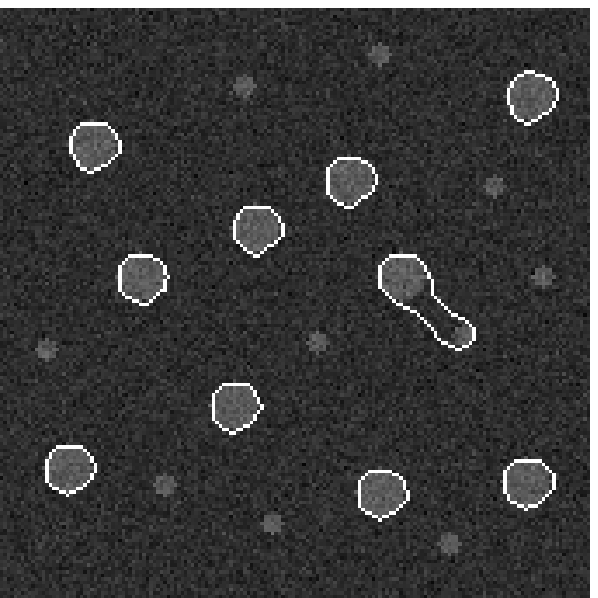} &
            \includegraphics[width=0.32\textwidth]{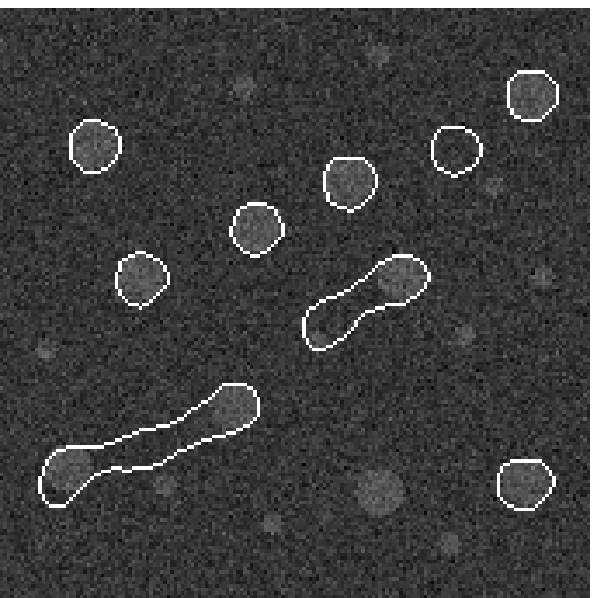} \\
            $(0.65, 0.098, 0.37, 0.099, 1.4, 1.4, 2.5)$ &
            $(0.60, 0.11, 0.43, 0.11, 0.51, 0.51, 0.89)$
       \end{tabular}
    \end{center}
    \caption{One of the synthesized images, with six different levels
    of added white Gaussian noise. Reading from left to right, top to
    bottom, the image variance to noise power ratios are $20$, $15$,
    $10$, $5$, $0$, $-5$ dB. Parameter values in the form $(\muin,
    \sigmain, \muout, \sigmaout, \lambdac, \alphac, \betac)$ are
    shown under the six images. The parameters $\dmin$ and $\ro$ were
    fixed to $8$ throughout.} \label{fig:synth_noise}
\end{figure}

The results obtained on the noisy versions of one of the fifty images
are shown in figure~\ref{fig:synth_noise}. Table~\ref{tab:noise}
shows the proportion of false negative and false positive circle
detections with respect to the total number of potentially correctly
detectable circles ($500 = 50 \times 10$), as well as the proportion
of `joined circles', when two circles are grouped together (an
example can be seen in the bottom right image of
figure~\ref{fig:synth_noise}). Detections of one of the smaller
circles (which only occurred a few times even at the highest noise
level) were counted as false positives. The method is very robust
with respect to all but the highest levels of noise. The first errors
occur at $5$ dB, where there is a $2\%$ false positive rate. At $0$
dB, the error rate is $\sim 10\%$, \ie one of the ten circles in each
image was misidentified on average. At $-5$ dB, the total error rate
increases to $\sim 30\%$, rendering the method not very useful.

Note that the principal error modes of the model are false positives and
joined circles. There are good reasons why these two types of error
dominate. We will discuss them further in section~\ref{sec:conclusion}.
\begin{table}
    \begin{center}
        \begin{tabular}{|c||c|c|c|}
            \hline
            noise (dB) &  FP (\%) & FN (\%)  & J (\%) \\
            \hline
            20 & 0 & 0 & 0 \\
            \hline
            15 & 0 & 0 & 0 \\
            \hline
            10 & 0 & 0 & 0 \\
            \hline
            5 & 2 & 0 & 0 \\
            \hline
            0 & 6.4 & 4 & 0 \\
            \hline
            -5 & 27.6 & 3.6 & 23 \\
            \hline
        \end{tabular}
    \end{center}
    \caption{Results on synthetic noisy images. FP, FN, J: percentages of
    false positive, false negative, and joined circle detections
    respectively, with respect to the potential total number of correct
    detections.} \label{tab:noise}
\end{table}

\subsection{Circle separation: comparison to classical active contours}
\label{sec:circleseparation}

In a final experiment, we simulated one of the most important causes of
error in tree crown extraction, and examined the response of classical
active contour and HOAC models to this situation. The errors, which involve
joined circles similar to those found in the previous experiment, are
caused by the fact that in many cases nearby tree crowns in an image are
connected by regions of significant intensity with significant gradient
with respect to the background, thus forming a dumbbell shape. Calling the
bulbous extremities, the `bells', and the join between them, the `bar', the
situation arises when the bells are brighter than the bar, while the bar is
in turn brighter than the background, and most importantly, the gradient
between the background and the bar is greater than that between the bar and
the bells.

\begin{figure}[h]
    \begin{center}
        \framebox{\includegraphics[width=0.55\textwidth]{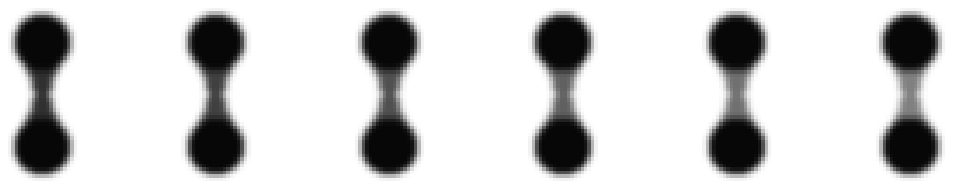}}
        \framebox{\includegraphics[width=0.55\textwidth]{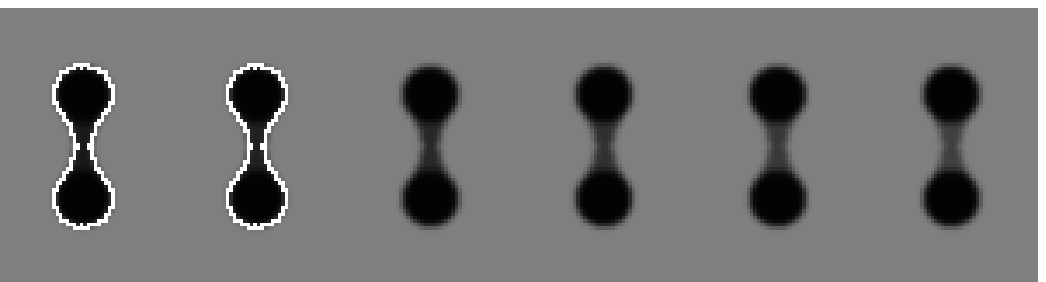}}
        \framebox{\includegraphics[width=0.55\textwidth]{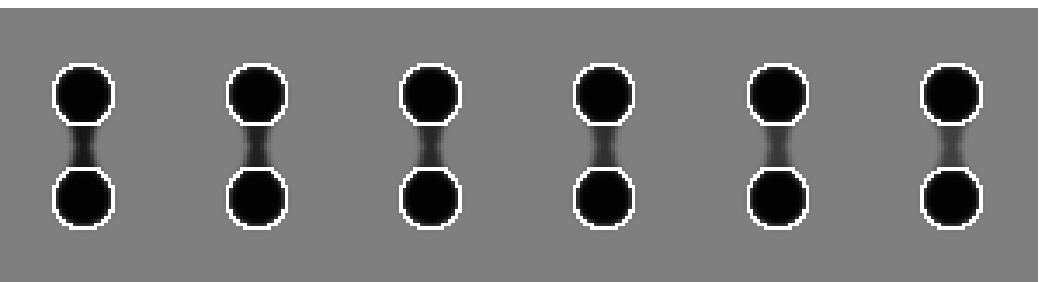}}
    \end{center}
    \caption{Results on circle separation comparing the HOAC `gas of
    circles' model to the classical active contour model. Top: the original images. The
    intensity of the bar takes values equally spaced between $48$ and
    $128$ from left to right; the background is $255$; the bells are
    $0$. In the middle: the best results obtained using the classical active contour model
    $(8, 1, 1)$. Either the circles are not separated or the region
    vanishes. Bottom: the results using the HOAC `gas of circles'
    model $(2, 1, 5, 4.0, 8, 8)$. All the circles are segmented correctly.}
    \label{fig:synthgrad}
\end{figure}

The first row of figure~\ref{fig:synthgrad} shows a sequence of bells
connected by bars. The intensity of the bar varies along the sequence,
resulting in different gradient values. We applied the classical active
contour and HOAC `gas of circles' models to these images.

The middle row of figure~\ref{fig:synthgrad} shows the best results
obtained using the classical active contour model. The model was either
unable to separate the individual circles, or the region completely
vanished. The intuition is that if there is insufficient gradient to stop
the region at the sides of the bar, then there will also be insufficient
gradient to stop the region at the boundary between the bar and the bells,
so that the region will vanish. On the other hand, if there is sufficient
gradient between the bar and the background to stop the region, the circles
will not be separated, and a `bridge' will remain between the two
circles.\footnote{`Bar' and `bell' refer to image properties; we use
`bridge' and `circle' to refer to the corresponding pieces of a
dumbbell-shaped region.}

The corresponding results using the HOAC `gas of circles' model can be seen
in the bottom row of figure~\ref{fig:synthgrad}. All the circles were
segmented correctly, independent of the gray level of the junction.
Encouraging as this is, it is not the whole story, as we indicated in
section~\ref{sec:noiseexperiments}. We make a further comment on this issue
in section~\ref{sec:conclusion}, which now follows.

\section{Conclusion}
\label{sec:conclusion}

Higher-order active contours allow the inclusion of sophisticated
prior information in active contour models. This information can
concern the relation between a region and the data, \ie the
likelihood $\pr{\imag | \reg, \knowledge}$, but more often it
concerns the prior probability $\pr{\reg | \knowledge}$ of a region,
or in other words, its `shape'. HOACs are particularly well adapted
to including shape information about regions for which the topology
is unknown {\em a priori}.

In this paper, we have shown via a stability analysis that a HOAC
energy can be constructed that describes a `gas of circles', that is,
it favours regions composed of an {\em a priori} unknown number of
circles of a certain radius, with short-range interactions amongst
them. The requirement that circles be stable, \ie local minima of the
energy, fixes one of the prior parameters and constrains another.

The `gas of circles' model has many potential uses in computer vision
and image processing. Combined with an appropriate likelihood, we
have applied it to the extraction of tree crowns from aerial images.
It performs better than simpler techniques, such as maximum
likelihood and standard active contours. In particular, it is better
able to separate trees that appear joined in the data than a
classical active contour model.

The model is not without its issues, however. The two most
significant are related to the principal error modes found in the
noise experiments of section~\ref{sec:noiseexperiments}: circles are
found where the data does not ostensibly support them (false
positives, a.k.a. `phantom' circles), and two circles may be joined
into a dumbbell shape and never separated. We discuss these in turn.

The first issue is that of `phantom' circles. Circles of radius
$\hatro$ are local minima of the prior energy. It is the effect of
the data that converts such configurations into global minima. Were
we able to find the global minimum of the energy, this would be fine.
However, the fact that gradient descent finds only a local minimum
can create problems in areas where the data does not support the
existence of circles. This is because a circle, once formed during
gradient descent, cannot disappear unless there is an image force
acting on it. We thus find that circles can appear and remain even
though there is no data to support them. Adding a large level of
noise exacerbates this problem, because random fluctuations may
encourage the appearance of circles as intermediate states during
gradient descent.

The second issue is that of joined circles, discussed in
section~\ref{sec:circleseparation}. Although the current HOAC model is
better able to separate circles than a classical active contour, it still
fails to do so in a number of cases, leaving a bridge between the circles.
The issue here is a delicate balance between the parameters, which must be
adjusted so that the sides of the bridge attract one another, thus breaking
the bridge, and so that nearby circles repel one another at close range, so
that the bridge does not re-form. Again, this is at least in part an
algorithmic issue. Even if the two separated circles have a lower energy
than the joined circles, separation may never be achieved due to a local
minimum caused by the bridge. Again, high levels of noise encourage this
behaviour by producing by chance image configurations that weakly support
the existence of a bridge.

We are currently working on solving both these problems through a
more detailed theoretical analysis of the energy, and in particular
the dependence of local minima on the parameters.

\appendix

\section{Details of stability computations} \label{AppendixA}

In this appendix, starting from the equation for the circle and the
expression for the radial perturbation in terms of its Fourier
coefficients,
    \begin{subequations}\label{eq:circleandperturb}
    \begin{align}
        \contour(t) & = \contouro(t) + \dcontour(t) = (r(t), \theta(t)) = (\ro(t) + \dr(t), \thetao(t))
        \label{eq:perturbedcontour} \\
        \intertext{where}
        \contouro(t) & = (\ro(t), \thetao(t)) =  (\ro, t) \label{eq:circle}\\
        \intertext{and}
        \dr(t) & = \sum_{k} a_{k} e^{i\ro kt}
        \eqcomma \label{eq:radius2}
    \end{align}
    \end{subequations}

\noi with $k \in \{m /\ro \st m \in \integers\}$, we give most of the steps
involved in reaching the expression,
equation~\eqref{eq:fullenergytosecondorder}, for the expansion to second
order of $\Eg$ around a circle.

The derivative of $\contour$ is given by
    \begin{subequations}\label{eq:derivatives}
    \begin{align}
        \dot{\theta}(t) & = 1 \label{eq:derivativetheta} \\
        \dot{r}(t) & = \dot{\dr}(t) = \sum_{k} a_{k} i \ro  k e^{i\ro kt}
        \eqstop \label{eq:derivativer}
    \end{align}
    \end{subequations}

\noi The tangent vector field is given by
    \begin{equation}
        \tanvec(t) = \dot{r}(t) \partial_{r} +
        \dot{\theta}(t)\partial_{\theta} \eqstop \label{eq:tanvec}
    \end{equation}

\noi We need the magnitude of this vector to second order. The metric in
polar coordinates is given by $ds^{2}=dr^{2}+r^{2} d\theta^{2}$, so we have
that $\abs{\tanvec(t)}^{2} = \dot{r}(t)^{2} + r(t)^{2}$ by
equation~\eqref{eq:derivativetheta}. Substituting from
equations~\eqref{eq:circleandperturb} and~\eqref{eq:derivativer} gives
    \begin{equation}
        \abs{\tanvec(t)}^{2}
        = \ro ^{2} + 2 \ro  \sum_{k} a_{k} e^{i \ro  k t} + \sum_{k, k'} a_{k}
        a_{k'} e^{i \ro  (k + k') t} (1 - \ro ^{2} k k')
        \eqstop\label{eq:tanvecsquared}
    \end{equation}

\noi Taking the square root, expanding it as $\sqrt{1+x} \approx 1 +
\frac{1}{2} x - \frac{1}{8} x^{2}$, and keeping terms to second order in
the $a_{k}$ then gives
    \begin{equation}
        \abs{\tanvec(t)}
        = \ro   \left\{1 + \sum_{k} \frac{a_{k}}{\ro} e^{i\ro kt} -
        \frac{1}{2} \sum_{k, k'} a_{k} a_{k'} k k' e^{i\ro (k + k')t}
        \right\} \eqstop\label{eq:tanvecexpansion}
    \end{equation}

\subsection{Length}

Using equation~\eqref{eq:tanvecexpansion}, the boundary length is then
given to second order by
\begin{eqnarray} \label{eq.app.l}
    \nonumber
    \length(\contour) = \int_{-\pi}^{\pi} dt\intspace \abs{\tanvec(t)}
    =  2\pi \ro  \left\{1+\frac{a_{0}}{\ro} + \frac{1}{2}\sum_{k} k^{2} \abs{a_{k}}^{2}
 \right\} \eqcomma
\end{eqnarray}

\noi where we have used the fact that
    \begin{align}
        \int_{-\pi}^{\pi}dt \isp e^{i\ro k t} = 2\pi\delta(k)\eqcomma\label{eq:fourierbasis}
    \end{align}

\noi and that $a_{-k} = a_{k}^{\ast}$, where $\ast$ indicates complex
conjugation, because $\dr$ is real.

\subsection{Area}

We can write the interior area of the region as
    \begin{equation}
        \area(\contour) = \int_{-\pi}^{\pi}d\theta\intspace
        \int_{0}^{r(\theta)} dr'\intspace r'
        = \int_{-\pi}^{\pi}d\theta\intspace \frac{1}{2} r^{2}(\theta)
        \eqstop  \non
    \end{equation}

\noi Thus, using equations~\eqref{eq:circleandperturb}, and again using
equation~\eqref{eq:fourierbasis} to integrate Fourier basis elements, we
have that
    \begin{align}
        \area(\contour) = \pi \ro ^{2} + 2\pi \ro  a_{0} + \pi \sum_{k}
        \abs{a_{k}}^{2} \eqstop \label{eq.app.a}
    \end{align}

\subsection{Quadratic energy} \label{AppendixA.quad}

To compute the expansion of the quadratic term in equation~\eqref{eq.q.1}
for $\Eg$, we need the expansions of $\tanvec(t)\cdot\tanvec(t')$ and
$\interactionfunction(R(t, t'))$.

\subsubsection{Inner product of tangent vectors}

The tangent vector is given by equation~\eqref{eq:tanvec}, but we must take
care as $\tanvec(t)$ and $\tanvec(t')$ live in different tangent spaces, at
$\contour(t)$ and $\contour(t')$ respectively. Since parallel transport
does not preserve the coordinate basis vectors $\partial_{r}$ and
$\partial_{\theta}$, it will change the components of $\tanvec(t')$, say,
when we parallel transport it to the tangent space at $\contour(t)$. It is
easiest to convert the tangent vectors to the Euclidean coordinate basis,
    \begin{subequations}\non
    \begin{align}
        \partial_{r} & = \cos(\theta)\partial_{x} + \sin(\theta)\partial_{y}
        \non\\
        \partial_{\theta} & = -r\sin(\theta)\partial_{x} +
        r\cos(\theta)\partial_{y} \eqcomma \non
    \end{align}
    \end{subequations}

\noi as these basis vectors are preserved by parallel transport. Doing so,
and then taking the inner product gives
    \begin{multline}
        \tanvec \cdot \tanvec' =
        \cos(\theta' - \theta )  [\ro ^{2} + \ro  \dr
        + \ro  \dr' + \dr \dr' + \dot{\dr} \dot{\dr}'] \\
        + \sin(\theta' - \theta )  [\ro  \dot{\dr}' - \ro  \dot{\dr} + \dr \dot{\dr}'
        - \dot{\dr} \dr'] \eqstop \non
    \end{multline}

\noi where unprimed quantities are evaluated at $t$ and primed quantities
at $t'$. Note that when $t = t'$, the expression reduces to
equation~\eqref{eq:tanvecsquared}.

\subsubsection{Distance between two points}

The squared distance between $\contour(t')$ and $\contour(t)$ is given by
    \begin{align}
        \abs{\contour(t') - \contour(t)}^{2} & =(x(t') - x(t))^{2} + (y(t') -
        y(t))^{2}\non\\
        & =[(\ro  + \dr')  \cos(\theta ') - (\ro  + \dr)  \cos(\theta)]^{2}
        + [(\ro  + \dr') \sin(\theta ') - (\ro  + \dr) \sin(\theta)]^{2} \eqcomma\non
    \end{align}

\noi which after expansion gives
    \begin{equation}
        \abs{\contour(t') - \contour(t)}^{2} =
        2 \ro ^{2}(1 - \cos(\Dtheta))
        \biggl\{
            1+ \frac{1}{\ro}(\dr + \dr')+ \frac{\dr^{2} + \dr'^{2} - 2 \cos(\Dtheta) \dr \dr'}{2 \ro^{2}(1 - \cos(\Dtheta))}
        \biggr\}
        \eqcomma\non
    \end{equation}

\noi where $\Dtheta = \theta' - \theta = t'- t$. Expanding $\sqrt{1+x}
\approx 1 + \frac{1}{2} x - \frac{1}{8} x^{2}$ to second order and
collecting terms, we then find
    \begin{equation}
        R(t, t') = \abs{\contour(t') - \contour(t)} =
        2 \ro \abs{\sin(\Dtheta/2)}
        + \abs{\sin(\Dtheta/2)}(\dr + \dr')
        + \frac{A(\Dtheta)}{4\ro}
        (\dr - \dr')^{2} \eqcomma\label{eq:pointdistancesecondorder}
    \end{equation}

\noi where $A(z)= \left( \frac{\cos^{2} \left( \frac{z}{2} \right)}{\left|
\sin \frac{z}{2} \right|} \right)$.

\subsubsection{Interaction function}

Expanding $\interactionfunction(z)$ in a Taylor series to second order, and
then substituting $R(t, t')$ for $z$ using the approximation in
equation~\eqref{eq:pointdistancesecondorder}, and keeping only terms up to
second order in $\dcontour$ then gives
    \begin{multline}
        \interactionfunction(R(t, t')) =
        \interactionfunction(X_{0})
        + \bigabs{\sin\frac{\Dtheta}{2}}\interactionfunction'(X_{0}) (\dr + \dr') \\
        + \frac{1}{4\ro} A(\Dtheta) \interactionfunction'(X_{0}) (\dr - \dr')^{2}
        + \frac{1}{2} \sin^{2}\Bigl(\frac{\Dtheta}{2}\Bigr)
        \interactionfunction''(X_{0}) (\dr + \dr')^{2} \eqcomma
    \end{multline}

\noi where $X_{0}= 2 \ro\abs{\sin(\Dtheta/2)}$.

\subsubsection{Combining terms}

Now let $G(t, t')= \tanvec(t) \cdot \tanvec(t')\interactionfunction(R(t,
t'))$. Combining the expressions already derived, we have
    \begin{align}
        & G(t, t') = \non\\
        & \underbrace{\ro ^{2} \cos(\Dtheta) \interactionfunction(X_{0})}_{F_{00}, \text{ even}} \non\\
        & + (\dr + \dr')
        \underbrace {\ro  \cos(\Dtheta) \left\{\interactionfunction(X_{0})
        + \ro  \left\abs{\sin \frac{\Dtheta}{2} \right} \interactionfunction'(X_{0}) \right \}}_{F_{10}, \text{ even}}\non\\
        & + (\dot{\dr}' - \dot{\dr})
        \underbrace{\ro  \sin(\Dtheta) \interactionfunction(X_{0})}_{F_{11}, \text{ odd}}\non\\
        & + (\dr^{2} + \dr'^{2})
        \underbrace{\ro  \cos(\Dtheta) \left\{\frac{1}{4} A(\Dtheta) \interactionfunction'(X_{0})
        + \frac{1}{2}\ro  \sin^{2} \Bigl(\frac{\Dtheta}{2}\Bigr) \interactionfunction''(X_{0})
        + \bigabs{\sin \frac{\Dtheta}{2}} \interactionfunction'(X_{0})\right\}}_{F_{20}, \text{ even}}\non\\
        & + (\dr \dr')
        \underbrace{\cos(\Dtheta) \left\{\interactionfunction(X_{0})
        + 2 \ro  \bigabs{\sin \frac{\Dtheta}{2}} \interactionfunction'(X_{0})
        - \frac{1}{2} \ro  A(\Dtheta) \interactionfunction'(X_{0})
        + \ro ^{2} \sin^{2} \Bigl(\frac{\Dtheta}{2}\Bigr) \interactionfunction''(X_{0})\right\}}_{F_{21}, \text{ even}}\non\\
        & + (\dr' \dot{\dr}' - \dr \dot{\dr})
        \underbrace{\ro  \bigabs{\sin \frac{\Dtheta}{2}} \sin(\Dtheta) \interactionfunction'(X_{0})}_{F_{22}, \text{ odd}}\non\\
        & + (\dr \dot{\dr}' - \dr' \dot{\dr})
        \underbrace{\sin(\Dtheta) \left\{\interactionfunction(X_{0})
        + \ro  \bigabs{\sin \frac{\Dtheta}{2}} \interactionfunction'(X_{0})\right\}}_{F_{23}, \text{ odd}}\non\\
        & + (\dot{\dr} \dot{\dr}')
        \underbrace{\cos(\Dtheta)\interactionfunction(X_{0})}_{F_{24}, \text{ even}} \eqstop\non
    \end{align}

\noi where we have introduced the notation $F_{00}\ldots F_{24}$ for the
functions appearing in the terms of $G$, and `odd' and `even' refer to
parity under exchange of $t$ and $t'$. Note that the $F$ are functionals of
$\interactionfunction$, and functions of $\ro$ and $t' - t$ (but not $t$
and $t'$ separately). Note also that each line, and hence $G$, is symmetric
in $t$ and $t'$.

The integral in the quadratic energy term is now given by
$\iint_{-\pi}^{\pi} dt\isp dt'\isp G(t, t')$. We can now substitute the
expressions for $\dr$ and $\dot{\dr}$ in terms of their Fourier
coefficients, $\dr(t)= \sum_{k} a_{k} e^{i \ro  k t}$ and $\dot{\dr}(t)=
\sum_{k} a_{k} i \ro  k e^{i \ro k t}$. Due to the dependence of the $F$ on
$t - t'$ only, the resulting integrals can be reduced, via a change of
variables $p = t' - t$, to integrals over $p$. We note that in the terms
involving $F_{10}$, $F_{11}$, $F_{20}$, $F_{22}$, and $F_{23}$, the
presence of the symmetric or antisymmetric factors in $\dr$ and $\dr'$
simply leads to a doubling of the value of the integral for one of the
terms in these factors, due to the corresponding symmetry or antisymmetry
of the $F$ functions. For example,
    \begin{equation}
        \iint_{-\pi}^{\pi} dt\isp dt'\isp  (\dr \dot{\dr}' - \dot{\dr} \dr')  \isp F_{23}(t' - t) =
        2 \iint_{-\pi}^{\pi} dt\isp dt'\isp  \dr \dot{\dr}'  \isp F_{23}(t' - t)
        \eqstop\non
    \end{equation}

\noi We therefore only need to evaluate one of these integrals for the
relevant terms. Below we list the calculations for all the $F$ integrals
for completeness:
    \begin{align}
        \iint_{-\pi}^{\pi} dt\isp dt'\isp \isp F_{00}(t' - t) & =
        \iint_{-\pi}^{\pi} dp\isp dt'\isp \isp F_{00}(p) \non\\
        & = 2\pi\int_{-\pi}^{\pi} dp\isp \isp F_{00}(p) \eqcomma\non
    \end{align}

\noi which survives because $F_{00}$ is symmetric;

    \begin{align}
        \iint_{-\pi}^{\pi} dt\isp dt'\isp  \dr(t) \isp F_{10}(t' - t) & =
        \iint_{-\pi}^{\pi} dt\isp dt'\isp \sum_{k} a_{k} e^{i \ro  k t} \isp F_{10}(t' - t) \non\\
        & = \sum_{k} a_{k} \iint_{-\pi}^{\pi} dp\isp dt'\isp e^{i \ro  k (-p+t')} \isp F_{10}(p) \non\\
        & = \sum_{k} a_{k} \int_{-\pi}^{\pi} dt'\isp e^{i \ro  k t'} \int_{-\pi}^{\pi} dp\isp e^{-i \ro  k p} \isp F_{10}(p) \non\\
        & = \sum_{k} a_{k} 2\pi \delta(k) \int_{-\pi}^{\pi} dp\isp e^{-i \ro  k p} \isp F_{10}(p) \non\\
        & = 2\pi a_{0} \int_{-\pi}^{\pi} dp\isp \isp F_{10}(p) \eqcomma\non
    \end{align}

\noi which survives because $F_{10}$ is symmetric;

    \begin{align}
        \iint_{-\pi}^{\pi} dt\isp dt'\isp \dot{\dr}(t) \isp F_{11}(t' - t) & =
        \iint_{-\pi}^{\pi} dt\isp dt'\isp \sum_{k} a_{k} i \ro k  e^{i \ro  k t} \isp F_{11}(t' - t) \non\\
        & = \sum_{k} a_{k} i \ro k  \iint_{-\pi}^{\pi} dp\isp dt'\isp  e^{i \ro  k (-p+t')} \isp F_{11}(p) \non\\
        & = \sum_{k} a_{k} i \ro k  \int_{-\pi}^{\pi} dt'\isp e^{i \ro  k t'} \int_{-\pi}^{\pi} dp\isp e^{-i \ro  k p} \isp F_{11}(p) \non\\
        & = \sum_{k} a_{k} i \ro k  2\pi \delta(k) \int_{-\pi}^{\pi} dp\isp e^{-i \ro  k p} \isp F_{11}(p) \non\\
        & = 0 \eqsemi\non
    \end{align}

    \begin{align}
        \iint_{-\pi}^{\pi} dt\isp dt'\isp \dr^{2}(t) \isp F_{20}(t' - t) & =
        \iint_{-\pi}^{\pi} dt\isp dt'\isp \sum_{k} \sum_{k'} a_{k} a_{k'} e^{i \ro  (k + k') t} \isp F_{20}(t' - t) \non\\
        & = \sum_{k} \sum_{k'} a_{k} a_{k'} \iint_{-\pi}^{\pi} dp\isp dt'\isp  e^{i \ro  (k + k') (-p + t')} \isp F_{20}(p) \non\\
        & = \sum_{k} \sum_{k'} a_{k} a_{k'} \int_{-\pi}^{\pi} dt'\isp e^{i \ro  (k + k')t'} \int_{-\pi}^{\pi} dp\isp  e^{-i \ro  (k + k')p} \isp F_{20}(p) \non\\
        & = \sum_{k} \sum_{k'} a_{k} a_{k'} 2\pi \delta(k+k') \int_{-\pi}^{\pi} dp\isp  e^{-i \ro  (k + k')p} \isp F_{20}(p) \non\\
        & = \sum_{k}  a_{k} a_{-k} 2\pi \int_{-\pi}^{\pi} dp\isp F_{20}(p) \non\\
        & = 2\pi \sum_{k}  \abs{a_{k}}^{2} \int_{-\pi}^{\pi} dp\isp F_{20}(p) \eqcomma\non
    \end{align}

\noi which survives because $F_{20}$ is symmetric;

    \begin{align}
        \iint_{-\pi}^{\pi} dt\isp dt'\isp  \dr(t) \dr(t')  \isp F_{21}(t' - t) & =
        \iint_{-\pi}^{\pi} dt\isp dt'\isp  \sum_{k} \sum_{k'} a_{k} a_{k'} e^{i \ro  (k t + k't')} \isp F_{21}(t' - t) \non\\
        & = \sum_{k} \sum_{k'} a_{k} a_{k'} \iint_{-\pi}^{\pi} dp\isp dt'\isp   e^{i \ro  k (-p + t')} e^{i \ro  k' t'} \isp F_{21}(p) \non\\
        & = \sum_{k} \sum_{k'} a_{k} a_{k'} \int_{-\pi}^{\pi} dt'\isp e^{i \ro  (k + k')t'} \int_{-\pi}^{\pi} dp\isp  e^{-i \ro  k p} \isp F_{21}(p) \non\\
        & = \sum_{k} \sum_{k'} a_{k} a_{k'} 2\pi \delta(k + k') \int_{-\pi}^{\pi} dp\isp   e^{-i \ro  k p} \isp F_{21}(p) \non\\
        & = \sum_{k}  a_{k} a_{-k} 2\pi \int_{-\pi}^{\pi} dp\isp  e^{-i \ro  k p} \isp F_{21}(p) \non\\
        & = 2\pi \sum_{k}  \abs{a_{k}}^{2} \int_{-\pi}^{\pi} dp\isp e^{-i \ro  k p} \isp F_{21}(p) \eqsemi\non
    \end{align}

    \begin{align}
        \iint_{-\pi}^{\pi} dt\isp dt'\isp  \dr(t) \dot{\dr}(t)  \isp F_{22}(t' - t) & =
        \iint_{-\pi}^{\pi} dt\isp dt'\isp \sum_{k} \sum_{k'} a_{k} a_{k'} i \ro  k e^{i \ro  (k+k') t} \isp F_{22}(t' - t) \non\\
        & = \sum_{k} \sum_{k'} a_{k} a_{k'} i \ro  k \iint_{-\pi}^{\pi} dp\isp dt'\isp  e^{i \ro  (k + k') (-p+t')} \isp F_{22}(p) \non\\
        & = \sum_{k} \sum_{k'} a_{k} a_{k'} i \ro  k \int_{-\pi}^{\pi} dt'\isp e^{i \ro  (k + k')t'} \int_{-\pi}^{\pi} dp\isp  e^{-i \ro  (k + k') p} \isp F_{22}(p) \non\\
        & = \sum_{k} \sum_{k'} a_{k} a_{k'} i \ro  k 2\pi \delta(k + k') \int_{-\pi}^{\pi} dp\isp  e^{-i \ro  (k + k') p} \isp F_{22}(p) \non\\
        & = 0 \eqcomma\non
    \end{align}

\noi because with $k + k' = 0$ from the delta function, the integral
becomes one over $F_{22}$ only, which vanishes due to the antisymmetry of
$F_{22}$;

    \begin{align}
        \iint_{-\pi}^{\pi} dt\isp dt'\isp  \dr(t) \dot{\dr}(t')  \isp F_{23}(t' - t) & =
        \iint_{-\pi}^{\pi} dt\isp dt'\isp  \sum_{k} \sum_{k'} a_{k} a_{k'} i \ro  k' e^{i \ro  (kt + k't')} \isp F_{23}(t' - t) \non\\
        & = \sum_{k} \sum_{k'} a_{k} a_{k'} i \ro  k' \iint_{-\pi}^{\pi} dp\isp dt'\isp  e^{i \ro  (k(-p + t') + k't')} \isp F_{23}(p) \non\\
        & = \sum_{k} \sum_{k'} a_{k} a_{k'} i \ro  k' \int_{-\pi}^{\pi} dt'\isp e^{i \ro  (k + k')t'} \int_{-\pi}^{\pi} dp\isp  e^{-i \ro k p} \isp F_{23}(p) \non\\
        & = \sum_{k} \sum_{k'} a_{k} a_{k'} i \ro  k' 2\pi \delta(k + k') \int_{-\pi}^{\pi} dp\isp  e^{-i \ro  k p} \isp F_{23}(p) \non\\
        & = - 2\pi \sum_{k} \abs{a_{k}}^{2} i \ro  k \int_{-\pi}^{\pi} dp\isp  e^{-i \ro  k p} \isp F_{23}(p) \eqsemi\non
    \end{align}

    \begin{align}
        \iint_{-\pi}^{\pi} dt\isp dt'\isp \dot{\dr}(t) \dot{\dr}(t')  \isp F_{24}(t' - t) & =
        \iint_{-\pi}^{\pi} dt\isp dt'\isp \sum_{k} \sum_{k'} a_{k} a_{k'} i^{2} \ro ^{2} k k' e^{i \ro  (kt+k't')} \isp F_{24}(t' - t) \non\\
        & = - \sum_{k} \sum_{k'} a_{k} a_{k'} \ro ^{2} k k' \iint_{-\pi}^{\pi} dp\isp dt'\isp e^{i \ro  (k(-p+t')+k't')} \isp F_{24}(p) \non\\
        & = - \sum_{k} \sum_{k'} a_{k} a_{k'} \ro ^{2} k k' \int_{-\pi}^{\pi} dt'\isp e^{i \ro  (k+k')t'} \int_{-\pi}^{\pi} dp\isp e^{-i \ro  k p} \isp F_{24}(p) \non\\
        & = - \sum_{k} \sum_{k'} a_{k} a_{k'} \ro ^{2} k k' 2\pi \delta(k+k') \int_{-\pi}^{\pi} dp\isp e^{-i \ro  k p} \isp F_{24}(p) \non\\
        & = 2\pi \sum_{k} \abs{a_{k}}^{2} \ro ^{2} k^{2} \int_{-\pi}^{\pi} dp\isp e^{-i \ro  k p} \isp F_{24}(p)
        \eqstop\non
    \end{align}

Using these results then gives equation~\eqref{eq.f.q}, which in
combination with equations~\eqref{eq.f.l} and~\eqref{eq.f.a}, gives
equation~\eqref{eq:fullenergytosecondorder}.


\newpage

\begin{thebibliography}{32}
\providecommand{\natexlab}[1]{#1} \providecommand{\url}[1]{\texttt{#1}}
\expandafter\ifx\csname urlstyle\endcsname\relax
  \providecommand{\doi}[1]{doi: #1}\else
  \providecommand{\doi}{doi: \begingroup \urlstyle{rm}\Url}\fi

\bibitem[Andersen et~al.(2001)Andersen, Reutebuch, and Schreuder]{Andersen01}
H.E. Andersen, S.E. Reutebuch, and G.F. Schreuder.
\newblock Automated individual tree measurement through morphological analysis
  of a {LIDAR}-based canopy surface model.
\newblock In \emph{Proc. of the $1^{st}$ International Precision Forestry
  Symposium}, pages 11--21, Seattle, Washington, USA, June 2001.

\bibitem[Brandtberg and Walter(1998)]{Brandtberg98}
T.~Brandtberg and F.~Walter.
\newblock Automated delineation of individual tree crowns in high spatial
  resolution aerial images by multiple-scale analysis.
\newblock \emph{Machine Vision and Applications}, \penalty0 (2):\penalty0
  64--73, 1998.

\bibitem[Caselles et~al.(1993)Caselles, Catte, Coll, and Dibos]{Caselles93}
V.~Caselles, F.~Catte, T.~Coll, and F.~Dibos.
\newblock A geometric model for active contours.
\newblock \emph{Numerische Mathematik}, 66:\penalty0 1--31, 1993.

\bibitem[Caselles et~al.(1997)Caselles, Kimmel, and Sapiro]{Caselles97a}
V.~Caselles, R.~Kimmel, and G.~Sapiro.
\newblock Geodesic active contours.
\newblock \emph{International Journal of Computer Vision}, 22\penalty0
  (1):\penalty0 61--79, 1997.

\bibitem[Chen et~al.(2002)Chen, Tagare, Thiruvenkadam, Huang, Wilson, Gopinath,
  Briggs, and Geiser]{Chen02}
Y.~Chen, H.D. Tagare, S.~Thiruvenkadam, F.~Huang, D.~Wilson, K.S. Gopinath,
  R.W. Briggs, and E.A. Geiser.
\newblock Using prior shapes in geometric active contours in a variational
  framework.
\newblock \emph{International Journal of Computer Vision}, 50\penalty0
  (3):\penalty0 315--328, 2002.

\bibitem[Choquet-Bruhat et~al.(1996)Choquet-Bruhat, DeWitt-Morette, and
  Dillard-Bleick]{Choquet-Bruhat96}
Y.~Choquet-Bruhat, C.~DeWitt-Morette, and M.~Dillard-Bleick.
\newblock \emph{Analysis, Manifolds and Physics}.
\newblock Elsevier Science, Amsterdam, The Netherlands, 1996.

\bibitem[Cohen(1991)]{Cohen91}
L.D. Cohen.
\newblock On active contours and balloons.
\newblock \emph{CVGIP: Image Understanding}, 53:\penalty0 211--218, 1991.

\bibitem[Cohen and Kimmel(1997)]{Cohen97}
L.D. Cohen and R.~Kimmel.
\newblock Global minimum for active contour models: A minimal path approach.
\newblock \emph{International Journal of Computer Vision}, 24\penalty0
  (1):\penalty0 57--78, August 1997.

\bibitem[Cremers and Soatto(2003)]{Cremers03b}
D.~Cremers and S.~Soatto.
\newblock A pseudo-distance for shape priors in level set segmentation.
\newblock In \emph{Proceedings of the 2nd IEEE Workshop on Variational,
  Geometric and Level Set Methods}, pages 169--176, Nice, France, 2003.

\bibitem[Cremers et~al.(2002)Cremers, Tischh{\"a}user, Weickert, and
  Schn{\"o}rr]{Cremers02}
D.~Cremers, F.~Tischh{\"a}user, J.~Weickert, and C.~Schn{\"o}rr.
\newblock Diffusion snakes: Introducing statistical shape knowledge into the
  {M}umford-{S}hah functional.
\newblock \emph{International Journal of Computer Vision}, 50\penalty0
  (3):\penalty0 295--313, 2002.

\bibitem[Cremers et~al.(2003)Cremers, Kohlberger, and Schn{\"o}rr]{Cremers03a}
D.~Cremers, T.~Kohlberger, and C.~Schn{\"o}rr.
\newblock Shape statistics in kernel space for variational image segmentation.
\newblock \emph{Pattern Recognition}, 36\penalty0 (9):\penalty0 1929--1943,
  September 2003.

\bibitem[Cremers et~al.(2004)Cremers, Osher, and Soatto]{Cremers04b}
D.~Cremers, S.~Osher, and S.~Soatto.
\newblock Kernel density estimation and intrinsic alignment for
  knowledge-driven segmentation: Teaching level sets to walk.
\newblock In C.~Rasmussen \etal, editor, \emph{Proc. Patt. Rec.}, volume 3175
  of \emph{Lecture Notes in Computer Science}, pages 36--44, T{\"u}bingen,
  Germany, 2004.

\bibitem[Foulonneau et~al.(2003)Foulonneau, Charbonnier, and
  Heitz]{Foulonneau03}
A.~Foulonneau, P.~Charbonnier, and F.~Heitz.
\newblock Geometric shape priors for region-based active contours.
\newblock \emph{Proc. IEEE International Conference on Image Processing
  (ICIP)}, 3:\penalty0 413--416, 2003.

\bibitem[Geman and Geman(1984)]{Geman84}
S.~Geman and D.~Geman.
\newblock Stochastic relaxation, {G}ibbs distributions and the {B}ayesian
  restoration of images.
\newblock \emph{IEEE Transactions on Pattern Analysis and Machine
  Intelligence}, 6:\penalty0 721--741, 1984.

\bibitem[Gougeon(1998)]{Gougeon98a}
F.~A. Gougeon.
\newblock Automatic individual tree crown delineation using a valley-following
  algorithm and rule-based system.
\newblock In D.A. Hill and D.G. Leckie, editors, \emph{Proc. Int'l Forum on
  Automated Interpretation of High Spatial Resolution Digital Imagery for
  Forestry}, pages 11--23, Victoria, British Columbia, Canada, February 1998.

\bibitem[Gougeon(1995)]{Gougeon95a}
F.A. Gougeon.
\newblock A crown-following approach to the automatic delineation of individual
  tree crowns in high spatial resolution aerial images.
\newblock \emph{Canadian Journal of Remote Sensing, 21(3)}, pages 274--284,
  1995.

\bibitem[Grenander(1993)]{Grenander93}
U.~Grenander.
\newblock \emph{General Pattern Theory}.
\newblock Oxford University Press, Oxford, UK, 1993.

\bibitem[Kass et~al.(1988)Kass, Witkin, and Terzopoulos]{Kass88}
M.~Kass, A.~Witkin, and D.~Terzopoulos.
\newblock Snakes: Active contour models.
\newblock \emph{International Journal of Computer Vision}, 1\penalty0
  (4):\penalty0 321--331, 1988.

\bibitem[Larsen(1998)]{Larsen98}
M.~Larsen.
\newblock Finding an optimal match window for {S}pruce top detection based on
  an optical tree model.
\newblock In D.A. Hill and D.G. Leckie, editors, \emph{Proc. of the
  International Forum on Automated Interpretation of High Spatial Resolution
  Digital Imagery for Forestry}, pages 55--66, Victoria, British Columbia,
  Canada, February 1998.

\bibitem[Larsen(1999)]{Larsen99}
M.~Larsen.
\newblock Individual {T}ree {T}op {P}osition {E}stimation by {T}emplate
  {V}oting.
\newblock In \emph{Proc. of the Fourth International Airborne Remote Sensing
  Conference and Exhibition / $21^{st}$ Canadian Symposium on Remote Sensing},
  volume~2, pages 83--90, Ottawa, Ontario, June 1999.

\bibitem[Leventon et~al.(2000)Leventon, Grimson, and Faugeras]{Leventon00}
M.E. Leventon, W.E.L. Grimson, and O.~Faugeras.
\newblock Statistical shape influence in geodesic active contours.
\newblock In \emph{Proc. IEEE Computer Vision and Pattern Recognition (CVPR)},
  volume~1, pages 316--322, Hilton Head Island, South Carolina, USA, 2000.

\bibitem[Metaxas(1997)]{Metaxas97}
D.N. Metaxas.
\newblock \emph{Physics-based Deformable Models: Applications to Computer
  Vision, Graphics and Medical Imaging}.
\newblock Kluwer, 1997.

\bibitem[Miller and Younes(2002)]{Miller02}
M.~I. Miller and L.~Younes.
\newblock Group actions, homeomorphisms, and matching: A general framework.
\newblock \emph{International Journal of Computer Vision}, 41:\penalty0 61--84,
  2002.

\bibitem[Miller et~al.(1997)Miller, Grenander, O'Sullivan, and
  Snyder]{Miller97}
M.~I. Miller, U.~Grenander, J.~A. O'Sullivan, and D.~L. Snyder.
\newblock Automatic target recognition organized via jump-diffusion algorithms.
\newblock \emph{IEEE Transactions on Image Processing}, 6\penalty0
  (1):\penalty0 157--174, January 1997.

\bibitem[Osher and Sethian(1988)]{Osher88}
S.~Osher and J.~A. Sethian.
\newblock Fronts propagating with curvature dependent speed: Algorithms based
  on {H}amilton-{J}acobi formulations.
\newblock \emph{Journal of Computational Physics}, 79\penalty0 (1):\penalty0
  12--49, 1988.

\bibitem[Paragios and Rousson(2002)]{Paragios02a}
N.~Paragios and M.~Rousson.
\newblock Shape priors for level set representations.
\newblock In \emph{Proc. European Conference on Computer Vision (ECCV)}, pages
  78--92, Copenhagen, Denmark, 2002.

\bibitem[Perrin et~al.(2004)Perrin, Descombes, and Zerubia]{Perrin04}
G.~Perrin, X.~Descombes, and J.~Zerubia.
\newblock Tree crown extraction using marked point processes.
\newblock In \emph{Proc. European Signal Processing Conference (EUSIPCO)},
  Vienna, Austria, September 2004.

\bibitem[Perrin et~al.(2005)Perrin, Descombes, and Zerubia]{Perrin05}
G.~Perrin, X.~Descombes, and J.~Zerubia.
\newblock A marked point process model for tree crown extraction in
  plantations.
\newblock In \emph{Proc. IEEE International Conference on Image Processing
  (ICIP)}, Genova, Italy, September 2005.

\bibitem[Rochery et~al.(2003)Rochery, Jermyn, and Zerubia]{Rochery03b}
M.~Rochery, I.~H. Jermyn, and J.~Zerubia.
\newblock Higher order active contours and their application to the detection
  of line networks in satellite imagery.
\newblock In \emph{Proc. IEEE Workshop Variational, Geometric and Level Set
  Methods in Computer Vision}, at ICCV, Nice, France, October 2003.

\bibitem[Rochery et~al.(2005)Rochery, Jermyn, and Zerubia]{Rochery05d}
M.~Rochery, I.~H. Jermyn, and J.~Zerubia.
\newblock Higher order active contours.
\newblock Research Report 5656, {INRIA}, France, August 2005.

\bibitem[Rochery et~al.(2006)Rochery, Jermyn, and Zerubia]{Rochery06}
M.~Rochery, I.~H. Jermyn, and J.~Zerubia.
\newblock Higher-order active contours.
\newblock \emph{International Journal of Computer Vision}, 69\penalty0
  (1):\penalty0 27--42, 2006.
\newblock URL \url{http://dx.doi.org/10.1007/s11263-006-6851-y}.

\bibitem[Sundaramoorthi and Yezzi(2005)]{Sundaramoorthi05}
G.~Sundaramoorthi and A.~Yezzi.
\newblock More-than-topology-preserving flows for active contours and polygons.
\newblock In \emph{Proc. IEEE International Conference on Computer Vision
  (ICCV)}, pages 1276--1283, Beijing, China, 2005.

\end{thebibliography}
\end{document}